\documentclass{article}

\usepackage[preprint]{neurips_2026}
\usepackage[utf8]{inputenc} % allow utf-8 input
\usepackage[T1]{fontenc}    % use 8-bit T1 fonts
\usepackage{hyperref}       % hyperlinks
\usepackage{url}            % simple URL typesetting
\usepackage{booktabs}       % professional-quality tables
\usepackage{amsfonts}       % blackboard math symbols
\usepackage{nicefrac}       % compact symbols for 1/2, etc.
\usepackage{microtype}      % microtypography
\usepackage{xcolor}         % colors

\newcommand{\eg}{\textit{e.g.}}

\newcommand{\ie}{\textit{i.e.}}

\newcommand{\methodname}{vesselFM-CT}
\newcommand{\Methodname}{VesselFM-CT}
\newcommand{\lossname}{TubeLoss}
\newcommand{\metricname}{TubeDice}

\usepackage{graphicx}
\usepackage{pifont}
\usepackage{amsmath}
\usepackage{wrapfig} % check if allowed
\usepackage{multirow}

\usepackage{algorithm}
\usepackage{algpseudocode}

\algrenewcommand{\algorithmiccomment}[1]{\hfill// #1}
\title{vesselFM-CT: Segmenting \textit{All} Blood Vessels in CT Images for System-Level Cardiovascular Analysis}

\author{%
  Bastian Wittmann \quad Chinmay Prabhakar \quad Suprosanna Shit \quad Bjoern Menze \vspace{0.5em}\\
  Department of Quantitative Biomedicine, University of Zurich, Zurich, Switzerland \\
  \texttt{\{bastian.wittmann, bjoern.menze\}@uzh.ch} \vspace{0.5em}\\
  \url{https://github.com/bwittmann/vesselFM-CT}
}

\begin{document}

\maketitle

\begin{abstract}
% 3D blood vessel segmentation presents one of largest unsolved problems in the field of medical image analysis.
The vascular network in the human body is characterized by blood vessels exhibiting drastic structural variations in radius, length, topological properties, and branching patterns. This heterogeneity, together with location-specific anatomical background variations, poses a significant challenge for robust, large-scale analysis of the entire cardiovascular system. As a result, most research has focused on narrow, isolated segments of the vascular network. 
While such targeted studies provide valuable insights, they inherently limit the ability to assess the systemic health and functional integrity of the vascular network as a whole.
In this work, we aim to bridge this gap to advance both clinical diagnostics and our fundamental understanding of vascular physiology. We propose the task of segmenting all vessels in CT images, ranging from the largest components of the cardiovascular system to even minuscule mesenteric vessels. To this end, we introduce \methodname{}, the first model capable of robustly segmenting all blood vessels in 3D CT images.
\Methodname{} is trained via an iterative, multi-step process and optimizes our proposed \lossname{} loss function, effectively addressing the inherent heterogeneity of the cardiovascular system.
We demonstrate that \methodname{} outperforms all baselines and enables automated, precise extraction of the cardiovascular system from CT images, thereby unlocking a wide range of clinical and technical perspectives, including automated disease classification and synthetic CT image generation.
\end{abstract}
\section{Introduction}
% Cardiovascular system: what does it do, why is it necessary?
The cardiovascular system supplies the human body with essential resources while facilitating waste removal, thermoregulation, and immune function. It ensures the proper functioning of all organ systems and consequently constitutes a fundamental and indispensable component of human physiology.
% Cardiovascular system: why does it need to be studied?
Despite this critical role, it is frequently affected by disorders including aneurysms, hypertension, stroke, or thrombosis, all of which contribute substantially to global morbidity and mortality~\cite{global2025global,WHO2025factsheet}.
% Cardiovascular system: how is does it relate to other things?
A deeper understanding of cardiovascular function and its interactions is therefore essential for the prevention and management of these diseases. 
Consequently, studying the cardiovascular system comprehensively across health and disease at a large scale is of central importance.

\begin{figure}[th]
\centering
\includegraphics[width=0.85\linewidth]{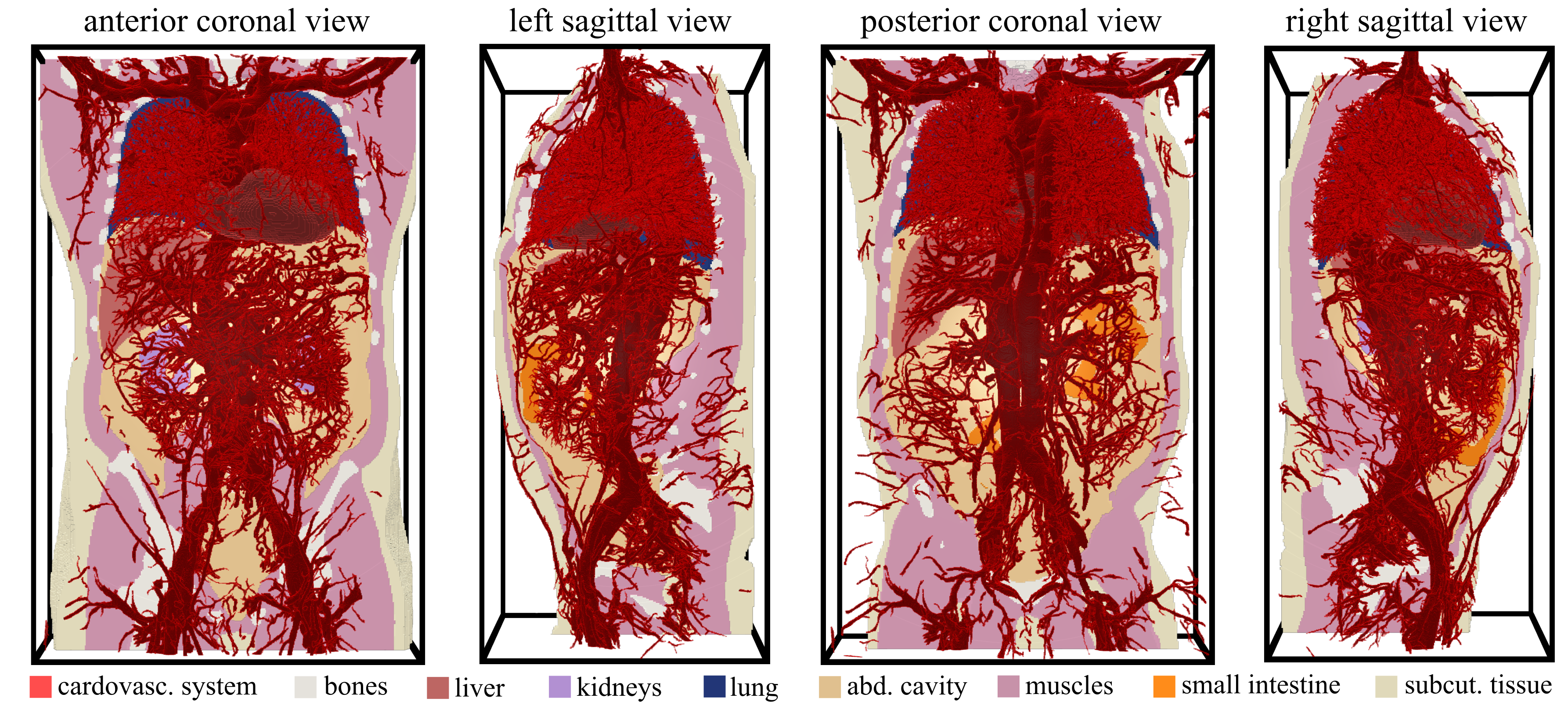}
\vspace{-0.5em}
\caption{
\Methodname{}'s prediction on a sample from the MSD dataset. For illustrative purposes, we also present the anatomical structures derived from the CADS-model. We find that \methodname{} segments the cardiovascular network in unprecedented detail across scales.
}
\label{fig:p1}
\end{figure}

% Deep learning for large scale analysis
Large-scale automated analysis is typically achieved by training deep neural networks (DNNs) in a supervised manner on manually annotated datasets of a specific target domain (\eg, CT or MRI), followed by inference on population-level datasets to extract information about structures of interest. Prior work has extensively applied DNNs to individual components of the cardiovascular system, including coronary arteries~\cite{li2022automatic,muscogiuri2025clinical}, pulmonary vessels~\cite{chu2025deep}, hepatic vessels~\cite{herold2025mri}, selected cerebral vessels~\cite{wei2024knowledge}, and the aorta~\cite{sieren2022automated}. 
% Why not systemic analysis
However, a comprehensive analysis of the entire cardiovascular system remains unexplored, leaving valuable anatomical and functional context unexploited.
Currently, automated segmentation of the entire cardiovascular system faces several roadblocks: 1) the difficulty of producing comprehensive, voxel-level annotations of the extensive and highly complex cardiovascular network; 2) the intrinsic heterogeneity of vasculature in size, shape, and branching patterns; and 3) the cardiovascular system's distribution across diverse anatomical regions with differing tissue contexts, which limits model generalization.
%Currently, automated segmentation of the entire cardiovascular system faces several road blocks: 1) Generating comprehensive, voxel-level segmentation masks for the entire human vascular network to train DNNs is practically almost infeasible due to its immense length and the substantial time and effort required to produce accurate, 3D-consistent annotations of the numerous tiny, tortuous, and highly branched vessels; 2) system-level vascular segmentation is inherently challenging due to the extreme complexity and heterogeneity of the vascular network, in which blood vessels vary widely in radius, length, topology, and hierarchical branching patterns; and 3) the cardiovascular system spans multiple anatomical regions, each with distinct surrounding tissues and anatomical contexts, which complicates generalization across the body.
% What do we do and how do we overcome these issues?
In this work, we aim at overcoming all of these roadblocks to ultimately enable complete, system-level analysis of the cardiovascular network extracted from CT images.

We propose the task of segmenting all blood vessels in CT images. To this end, we introduce \methodname{}, to our knowledge, the first model capable of robustly segmenting all blood vessels visible in CT images, ranging from, \eg, the aorta to minuscule vessels approaching the image resolution limit (see Fig.~\ref{fig:p1}). \Methodname{} is trained iteratively via a multi-step process, involving manual correction, simulated data, and multiple post-processing steps. We further introduce and optimize the \lossname{} loss function. \lossname{} is tailored to cope with the cardiovascular network's extreme heterogeneity and unique class imbalance challenges. Specifically, it introduces a very natural vessel-specific bias aiming at weighting vessel segments equally, irrespective of their radii. Additionally, we build upon \lossname{}'s intuition to propose a suitable measure of performance, the \metricname{} metric. \Methodname{} outperforms all baselines in segmentation fidelity and expressiveness, capturing a substantially more complete and detailed vascular network. Finally, we demonstrate \methodname{}'s clinical and technical perspectives through experiments on disease classification and synthetic image generation. We summarize our contributions as follows:

\begin{itemize}
    \item We introduce the task of segmenting all blood vessels from CT images, aiming to enable system-level vascular analysis and propose a strong baseline model in \methodname{}.
    \item We introduce a tailored loss and complementary metric in \lossname{} and \metricname{}, dynamically adjusting to the cardiovascular network's distinct form of intrinsic class imbalance.
    \item We demonstrate that \methodname{} enables comprehensive vascular analysis and opens new avenues for downstream applications, including disease classification and image generation.
\end{itemize}    
\section{Related Work}

\subsection{Cardiovascular Segmentation in CT Images}
In the past, most work focused on analyzing individual components of the cardiovascular system extracted from CT and CTA images~\cite{li2022automatic,chu2025deep,wei2024knowledge,sieren2022automated,yang2025benchmarkingcowtopcowchallenge}. Recently, however, the trend emerged of segmenting all relevant anatomical structures in CT (whole-body) images, including the cardiovascular system. TotalSegmentator~\cite{wasserthal2023totalsegmentator}, \eg, enables automated segmentation of over 100 anatomical structures, including core components of the cardiovascular system, and was recently extended to incorporate additional vascular structures~\cite{hinck2025automatic}. The CADS-model~\cite{xu2025cads} builds upon TotalSegmentator and consistently outperforms it in expressiveness (167 anatomical structures) and Dice scores. It exhibits exceptional robustness and generalization, enabled by training on 22,022 CT images with high-quality annotations, verified through automated quality control based on shape priors and neural implicit functions.
%, enabled by training on the CADS-dataset, a large multi-institutional collection comprising 22,022 CT images of varying spatial resolutions with high-quality annotations, verified through automated quality control based on shape priors and neural implicit functions.
BiomedParse~\cite{zhao2025foundation,zhao2025boltzmann} is designed to unify the tasks of biomedical segmentation, detection, and recognition. It is jointly trained on 6.8 million image-mask-text triplets from nine modalities. It supports holistic, text-guided analysis of biomedical images, with the authors demonstrating that BiomedParse is capable of segmenting all structures of interest. VoxTell~\cite{rokuss2025voxtell} outperforms BiomedParse on a wide variety of zero-shot segmentation tasks and represents a vision-language model for free-text-prompted universal medical image segmentation. Its training dataset contains over 62,000 medical images (CT, MRI, and PET), featuring an extensive array of vascular structures.\\
%, including arteries and veins of the anterior and posterior circulations, the carotid system, thoracic vessels, as well as blood vessels in the pancreas, liver, gonads, and kidneys. 
%The SAT (Segment Anything with Text) model~\cite{zhao2025large} has been designed to segment all structures in medical images from text prompts and relies on knowledge-enhanced representation learning for optimal vision-language alignment.
%Recently, Wittmann~\etal~\cite{wittmann2025vesselfm} proposed vesselFM, a foundation model capable of \textit{zero}-shot blood vessel segmentation in arbitrary 3D images. Even though vesselFM demonstrates promising initial predictions when applied directly to CT images, its strong tubular bias results in false positive tubular appearing structures, such as the intestines, ribs, and tendons. 
A fundamental limitation shared by these models is that they treat the cardiovascular system as a set of isolated anatomical structures from multiple data sources rather than as a continuous whole. As a result, they fail to capture its full complexity and are therefore unsuitable for system-level analysis.
% A fundamental limitation shared by these universal segmentation models is that the cardiovascular system is treated as a collection of selected and isolated anatomical structures derived from multiple data sources rather than as a continuum. As a consequence, all above introduced models fail to segment the cardiovascular system it its full complexity, rendering them, in contrast to \methodname{}, unsuitable for system-level analysis.

\subsection{Blood Vessel-Specific Loss Functions}
% In medical image segmentation, the loss function is typically formulated as a compound loss combining the Dice and cross-entropy loss, referred to as DiceCE in this manuscript. 
Given that blood vessels are of distinct structure, multiple loss functions tailored to their properties have been introduced. A prominent example is soft-clDice (centerline Dice)~\cite{shit2021cldice}, a loss function aiming to preserve connectivity in segmentation masks, derived from the topology-preserving clDice metric. The clDice metric explicitly reduces blood vessels to their centerlines (or skeletons) and evaluates their overlap with the segmentation masks via topology precision and sensitivity. Building on clDice, the cbDice (centerline boundary Dice)~\cite{shi2024centerline} loss introduces boundary-aware components and aims at coping with diameter imbalance by propagating the squared inverse radius onto the centerline during loss computation. The recently proposed Skeleton Recall loss~\cite{kirchhoff2024skeleton} operates on a tubed skeleton, which can be pre-computed offline in contrast to above-described differentiable skeleton-based methods, and provides a denser supervision signal. Subsequently, the tubed skeleton is utilized to compute a soft recall loss, resulting in state-of-the-art performance on multiple  datasets.\\
All these loss functions reduce vessels to centerlines, resulting in a simplified, heavily constrained objective that, in practice, must be combined with standard functions (\eg, Dice) to achieve competitive results. Our proposed \lossname{} takes a more natural approach by shifting from centerlines to vessel segments, while incorporating inductive biases elegantly through voxel-level weighting.
    
\section{Methodology}
%Recently, vesselFM~\cite{wittmann2025vesselfm} has shown state-of-the-art performance on the task of zero-shot universal blood vessel segmentation in arbitrary modalities. When directly applied to CT images, however, vesselFM struggles with false positives (FPs), fails to reliably segment vessels deviating from a perfectly tubular shape, and struggles with small-scale vessels (see Fig.~\ref{fig:training}). Given the importance of CT imaging in medicine, we address these limitations by proposing \methodname{}, a model capable of robustly and accurately segmenting all blood vessels visible in CT images ranging from the largest components of the cardiovascular system to vessels approaching the image resolution limit.
Given the importance of CT imaging in medicine, we propose \methodname{}, a model capable of robustly and accurately segmenting all blood vessels visible in CT images ranging from the largest components of the cardiovascular system to vessels approaching the image resolution limit. We tackle the challenge of generating comprehensive voxel-level segmentation masks covering the entire vascular network by training \methodname{} in an iterative manner involving manual correction, synthetic data, and engineered postprocessing steps. During training, we optimize our introduced \lossname{} to address the class imbalance inherent in the cardiovascular network, while simultaneously preserving accurate vessel boundaries and introducing resilience to incomplete annotations. We further utilize our proposed \metricname{} metric, derived from \lossname{}, to provide an accurate measure of performance for vessel segmentation tasks. 
\Methodname{}'s training process (see Section~\ref{sec:training}) is described together with our proposed loss and metric (see Section~\ref{sec:loss_metric}) below.

%on our curated \datasetname{} dataset (see Section~X), which represents the first dataset to provide comprehensive voxel-level annotations of all components of the cardiovascular system in CT images. To further cope with the unique challenges of the task of segmentation of the complete cardiovascular network, we train \methodname{} by optimizing our proposed \lossname{} and select parameters based on the \metricname{} metric (see Section~X). \methodname{}'s training scheme is describe together with our proposed loss and metric below.

\subsection{\Methodname{}'s Training Process}\label{sec:training}
\begin{figure*}[t!]
\centering
\includegraphics[width=\linewidth]{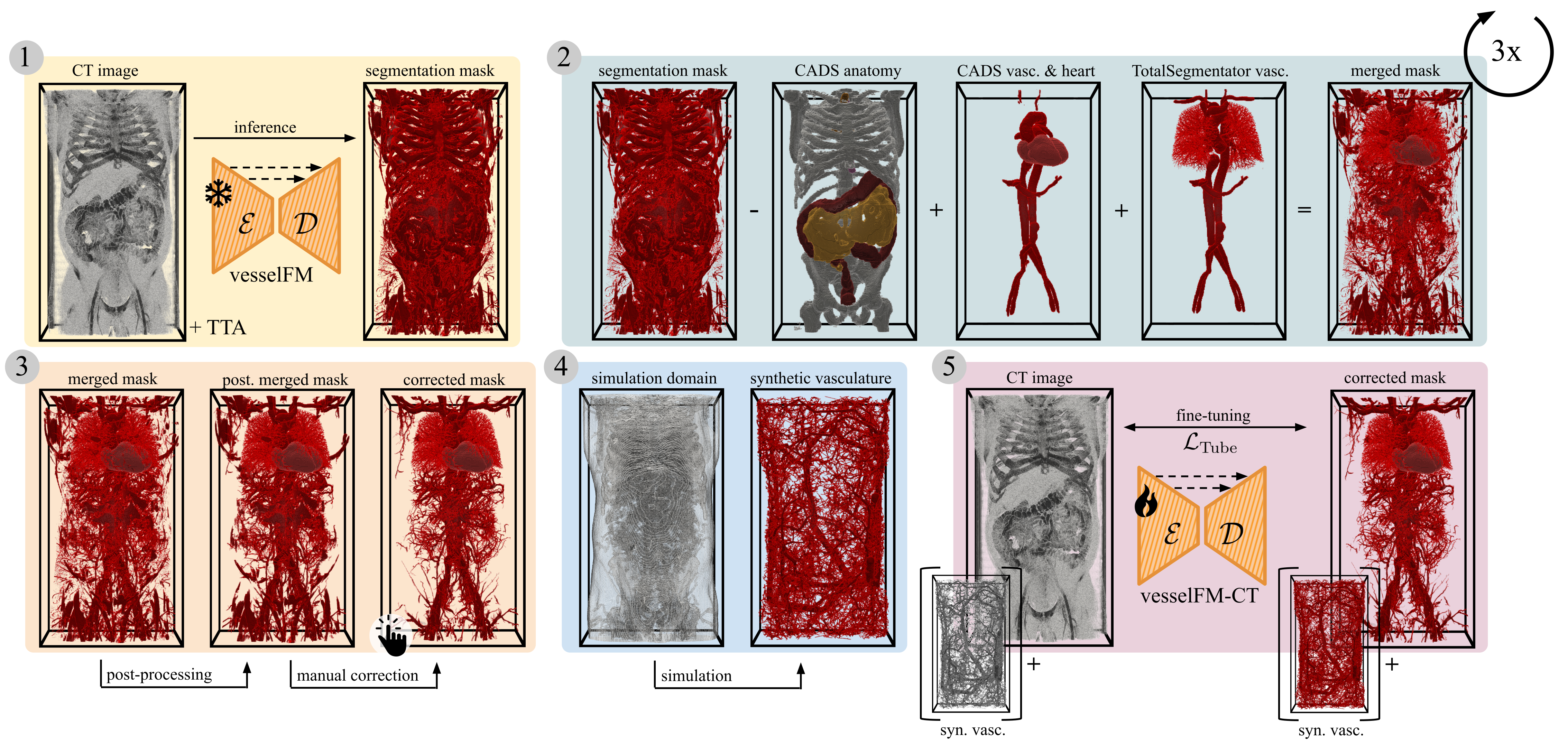}
\vspace{-1em}
\caption{
\Methodname{}’s multi-step training process. Step 1): We generate initial segmentation masks using vesselFM. Step 2): We leverage multi-anatomy segmentation models to suppress FPs and refine the vasculature. Step 3): We post-process the merged masks and manually correct them. Step 4): We generate synthetic vasculature. Step 5): We fine-tune \methodname{} on the corrected and synthetic data, optimizing our proposed \lossname{}. This process is iterated three times, progressively increasing the dataset size and quality. 
% The figure shows first-iteration results.
The mask in Step 1) excludes background FPs for clarity.
%\methodname{}'s multi-step training process: Step 1) We perform inference using vesselFM to generate initial segmentation masks; Step 2) We utilize label maps from multi-anatomy segmentation models to suppress FPs and refine vasculature; Step 3) We post-process the merged masks and perform manual correction; Step 4) We sample synthetic vasculature in tailored domains derived from the CT images; Step 5) We fine-tune vesselFM on the corrected masks combined with the synthetic vasculature using our novel \lossname{} loss function, resulting in \methodname{}. This process is iterated three times, progressively increasing the dataset size \methodname{}'s performance. It should be noted that the figure depicts results from the first iteration. The segmentation mask in Step 1) already excludes FPs in the image background for increased visibility.
}
\label{fig:training}
\end{figure*}

\Methodname{} is trained in a labor-intensive multi-step process, illustrated in Fig.~\ref{fig:training}. We build upon data from the Medical Segmentation Decathlon (MSD)~\cite{antonelli2022medical}, one of the most widely used datasets in medical image analysis. %with a permissive license.
Specifically, we build upon \texttt{Task03\_Liver}, given its relatively high voxel resolution and use of contrast agent. We resample all images to isotropic resolution to maintain the tubular shape of blood vessels, and solely include CT images with full torso coverage (shoulders to pelvis).
In the following, we describe each step in \methodname{}'s training process in detail.

\paragraph{Step 1):}
We build upon vesselFM~\cite{wittmann2025vesselfm}, a recently proposed foundation model capable of zero-shot blood vessel segmentation in 3D images of arbitrary modalities. Our early experiments indicate that applying vesselFM directly on CT images results in promising initial segmentation masks, capturing a large amount of tubular structures (see Fig.~\ref{fig:training}). To enhance robustness and overall segmentation performance at inference time, we introduce an expressive suite of test-time augmentations (TTA), consisting of aorta-based intensity alignment and axis flipping ($x$-, $y$-, and $z$-axis) at three different scales ($1\times$, $1.33\times$, $2\times$) with scale-specific prediction thresholds (0.5, 0.75, 0.75).

\paragraph{Step 2):}
Given vesselFM's strong tubular bias, the initial segmentation masks contain numerous tubular-appearing false positives (FPs) (\eg, intestines, ribs, tendons, muscles, urinary tract, bile ducts, and tubular components of the skeleton), as well as FPs caused by sharp contrast changes. We subsequently rely on predictions from the CADS-model to suppress them. To be specific, we exclude FPs originating from the esophagus, the trachea, the small intestine, the colon, the spine canal, bones, and the image background. Given that vesselFM further fails to reliably segment vessels deviating from a perfectly tubular shape and struggles with small-scale vessels, we additionally merge our refined predictions with vessels predicted by the CADS-model and TotalSegmentator (tasks \texttt{total} and \texttt{lung\_vessels}~\cite{poletti2022automated}), assigning all vessels to class 1. We exploit the redundancy between both models to enhance robustness, and solely add structures improving upon vesselFM's initial prediction. For completeness, we retain the heart from the CADS-model and assign it to class 2.

\paragraph{Step 3):}
The label map is post-processed via automated removal of small, unconnected components and manual correction. Manual correction focuses on adding missing vascular structures, refining boundaries, and eliminating remaining FPs. Manual correction is performed in ITK-SNAP~\cite{py06nimg} over the course of months, with special attention to ensure 3D voxel-level consistency of annotations. % in all three planes (sagittal, coronal, and transverse).

\paragraph{Step 4):}
Generating 3D-consistent annotations of minuscule vessels is a very challenging task. We find the task of erasing FPs to be significantly simpler and time-efficient compared to correcting FNs. Therefore, we generate synthetic vascular data to incorporate during fine-tuning, with the aim of increasing the model's sensitivity towards tubular structures (see~\ref{app:syn}). Our simulation relies on the \texttt{svv} package~\cite{sexton2025rapid,sexton_2025_svv}, which provides rapid synthetic vascular generation using constrained constructive optimization and coupled hemodynamic modeling.
We adjust its parameters to match small-scale vasculature and apply it to whole-body domains. To construct a realistic simulation domain matching a given CT image, we combine CADS-model labels with our generated label map, restricting vessel generation to the human body while preventing overlap with existing anatomical structures such as vessels and bones. The resulting vascular graph is transformed to a binary label map via a spherical kernel proportional to the vessel radii similar to~\cite{wittmann2024simulation}. To enhance realistic appearance, the label map is finally transformed via dilation, elastic deformation, erosion, and binary smoothing.

\paragraph{Step 5):}
We fine-tune \methodname{} using the updated label maps and randomly merge the synthetic vasculature into CT images and their corresponding label maps with a probability $p_\text{syn}$. While combining the synthetic vasculature with the label map is straight forward, merging it into the image requires additional consideration.
To this end, we first estimate the mean vessel intensity $\mu_{\text{vessel}}$ and assign it to the synthetic vasculature with an additional random intensity offset ($\Delta I \sim \mathcal{U}(-25\;\text{HU}, 100\;\text{HU})$). Subsequently, we add Gaussian noise ($\varepsilon \sim \mathcal{N}(0, \sigma^2), \text{where}\; \sigma \sim \mathcal{U}(10\;\text{HU}, 25\;\text{HU})$) onto the synthetic vasculature and finally merge it into the CT image via element-wise maximum fusion.
% by taking the element-wise maximum of the CT image's intensity and the synthetic vasculature's intensity. 
To focus solely on tubular structures, we exclude the heart from fine-tuning and optimize exclusively on class 1. 
During fine-tuning, we utilize our proposed \lossname{} (see Section~\ref{sec:loss_metric}) to cope with the inherent imbalance and missing annotations, maximizing the \metricname{} metric on the validation set.

We repeat our entire training process's steps three times, using the fine-tuned \methodname{} from Step 5) to perform inference in the subsequent Step 1), gradually increasing the dataset size from $n^{(1)} = 3$ to $n^{(2)} = 6$ and $n^{(3)} = 9$. We utilize a corrected sample from the first iteration for validation and manually correct three additional samples after the final iteration to represent our test set.
%Accordingly, \methodname{} is trained on eight images and tested on three images, while parameters were tuned on the validation sample.

%\datasetname{} consists of two subsets, the \goldstandard{} set and the \silverstandard{} set. While \gssmall{} samples are represented by our $n_3$ manually corrected images, we leverage the final, fine-tuned vesselFM to construct the \sssmall{}. Specifically, we perform inference on large-scale data, which is then thoroughly checked to ensure compliance with our quality criteria, without any manual intervention. Given that, we reserve \gssmall{} labels for testing and validation and \sssmall{} labels for training during benchmarking in Section X. Dataset statistics of the \gssmall{} and \sssmall{} are shown in Table~\ref{tab:dataset}, while samples from the \gssmall{} and \sssmall{} subsets are visualized in Fig. X.

\subsection{\lossname{} and \metricname{} Metric}\label{sec:loss_metric}
Segmenting the complete cardiovascular network bears unique class imbalance challenges. We consider class imbalance as a layered problem that can be condensed to \emph{extrinsic imbalance} and \emph{intrinsic imbalance}.
The extrinsic imbalance is straightforward and is defined as the imbalance between the vessel class and the background. Tube-like structures in a network, however, also vary drastically in their structural properties (\eg, radii). We, therefore, define the intrinsic imbalance as the imbalance in the network itself - \ie, imbalance between large and small vessels.
\begin{wrapfigure}{r}{0.45\textwidth}
\centering
\vspace{-0.5em}
\includegraphics[width=\linewidth]{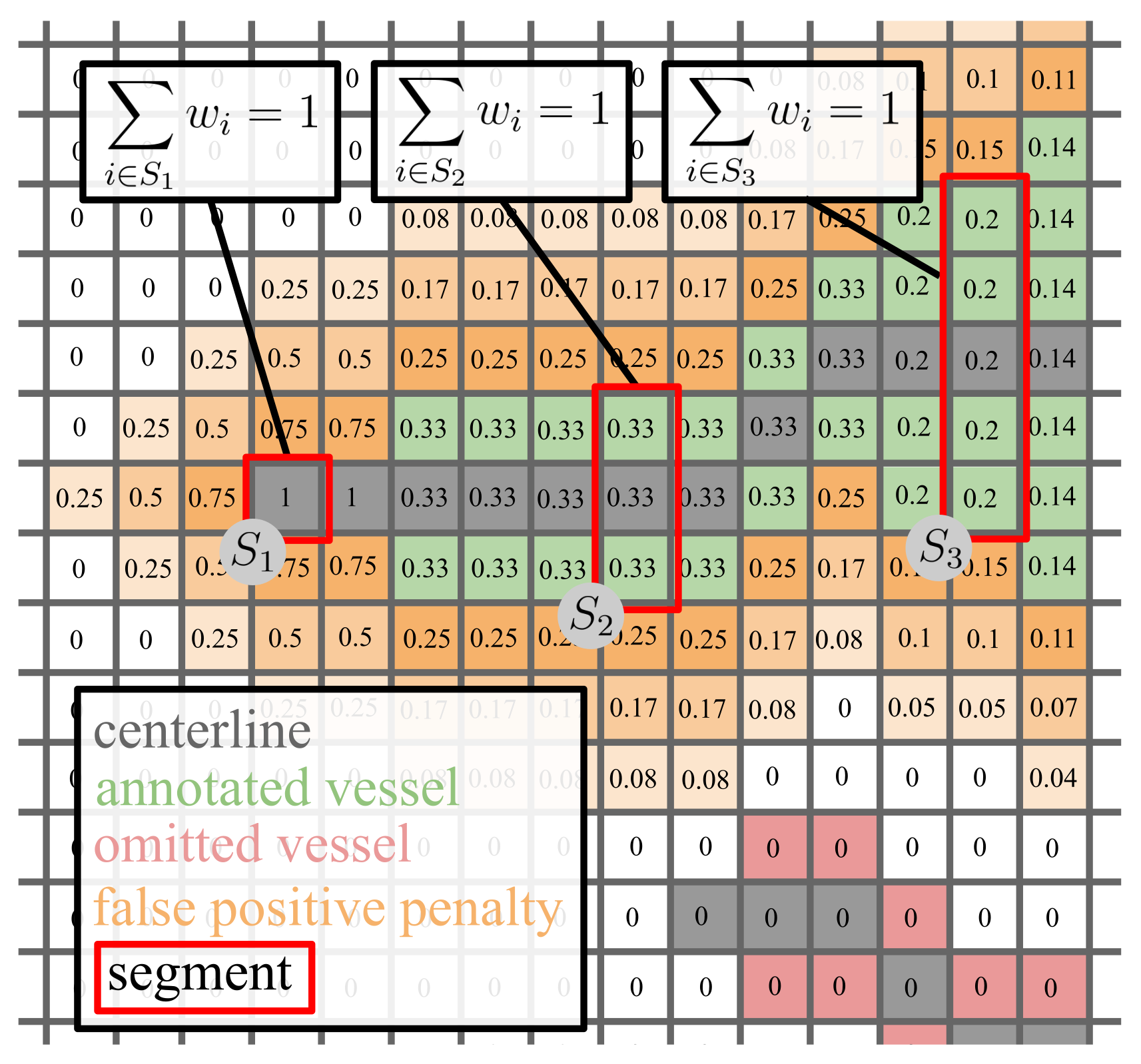}
\vspace{-1em}
\caption{
Unscaled weight map $w$ (shown in 2D for clarity). Segments (see $S_1$, $S_2$, and $S_3$) are equally weighted, while linear decaying FPP ($d_\text{FPP} = 3$) ensures accurate borders, and missing annotations are not over-penalized.
%
%, alleviating the inherent intrinsic class imbalance. Our weighting scheme does not over-penalize missing annotations (see omitted vessel), while the linear decaying FPP ($d_\text{FPP} = 3$ here)  ensures accurate vessel borders via high weights in their immediate surrounding.
}
\label{fig:w}
\end{wrapfigure}
Based on this intuition, we aim to construct a loss 
%tailored to the distinct form of class imbalance occurring in the cardiovascular network 
that dynamically adjusts to the distinct form of intrinsic class imbalance occurring in the cardiovascular network, weighting each segment in the network equally, agnostic to its cross-section (and radius).
% This behavior is especially of relevance when segmenting the cardiovascular system, given that it suffers heavily under intrinsic class imbalance.
To achieve this goal, we equip the commonly used compound loss combining the Dice loss and the cross-entropy (CE) loss with voxel-level weights $w$ derived from our above-described intuition.
%Given that the most commonly used metric for medical image segmentation is a compound loss combining the Dice loss and the Cross Entropy (CE) loss, \lossname{} equips both of them with a voxel-level weight map $w$ derived from our above describe intuition. 
%Although it is straight-forward to apply \lossname{} to multi-class segmentation tasks, we focus on binary blood vessel segmentation in this work. 
Accordingly, \lossname{} is defined as:

\begin{equation}
\mathcal{L}_{\text{Tube}} = \lambda_{\text{Dice}} \, \mathcal{L}_{\text{Dice}}^{(w)} + \lambda_{\text{CE}} \, \mathcal{L}_{\text{CE}}^{(w)},
\end{equation}

where $\mathcal{L}_{\text{Dice}}^{(w)}$ denotes the weighted Dice loss, $\mathcal{L}_{\text{CE}}^{(w)}$ the weighted binary CE loss, and $\lambda$ weighting coefficients (implementation details are mentioned in~\ref{app:loss}).

\begin{algorithm}[t]
\footnotesize
\caption{Generate voxel-level weight map $w$.}\label{alg:weights}
\begin{algorithmic}[1]
\Require 
GT $y \in \{0,1\}^{H\times W\times D}$,
FPP distance $d_\text{FPP} \in \mathbb{R}^+$,
scale factor $\alpha_{w} \in \mathbb{R}^+$,
radius cutoff  $r_\text{cutoff} \in \mathbb{R}^+$
\Ensure Voxel-level weight map $w \in \mathbb{R}^{H \times W \times D}$
\vspace{0.1cm}

\State $c \gets \texttt{skeletonize}(y)$ \Comment{Estimate blood vessel centerlines}
\State $r_c \gets \texttt{estimate\_radius\_for\_cl}(y, c)$ \Comment{Estimate radius of centerline}
\State $r \gets \texttt{propagate\_radius\_to\_mask}(y, r_c)$ \Comment{Assign radius of mask voxels to nearest centerline radius}
\State $r[r > r_\text{cutoff}] \gets r_\text{cutoff}$ \Comment{Clip radius}
\vspace{0.1cm}

\State $a \gets \texttt{estimate\_pixel\_area}(r)$ \Comment{Compute pixel area from radius}
\State $w \gets 1 / a$ \Comment{Model weights as inverse area}
\State $w \gets w + \texttt{linear\_decay}(d_\text{FPP}, w)$ \Comment{Generate and merge FPP into weight map}

\vspace{0.1cm}
\State $ w \gets (\alpha_w \cdot w) + 1$ \Comment{Scale and shift to generate final weight map $w$}
\State \Return $w$
\end{algorithmic}
\end{algorithm}

\begin{wrapfigure}{r}{0.28\textwidth}
\centering
\vspace{-1em}
\includegraphics[width=\linewidth]{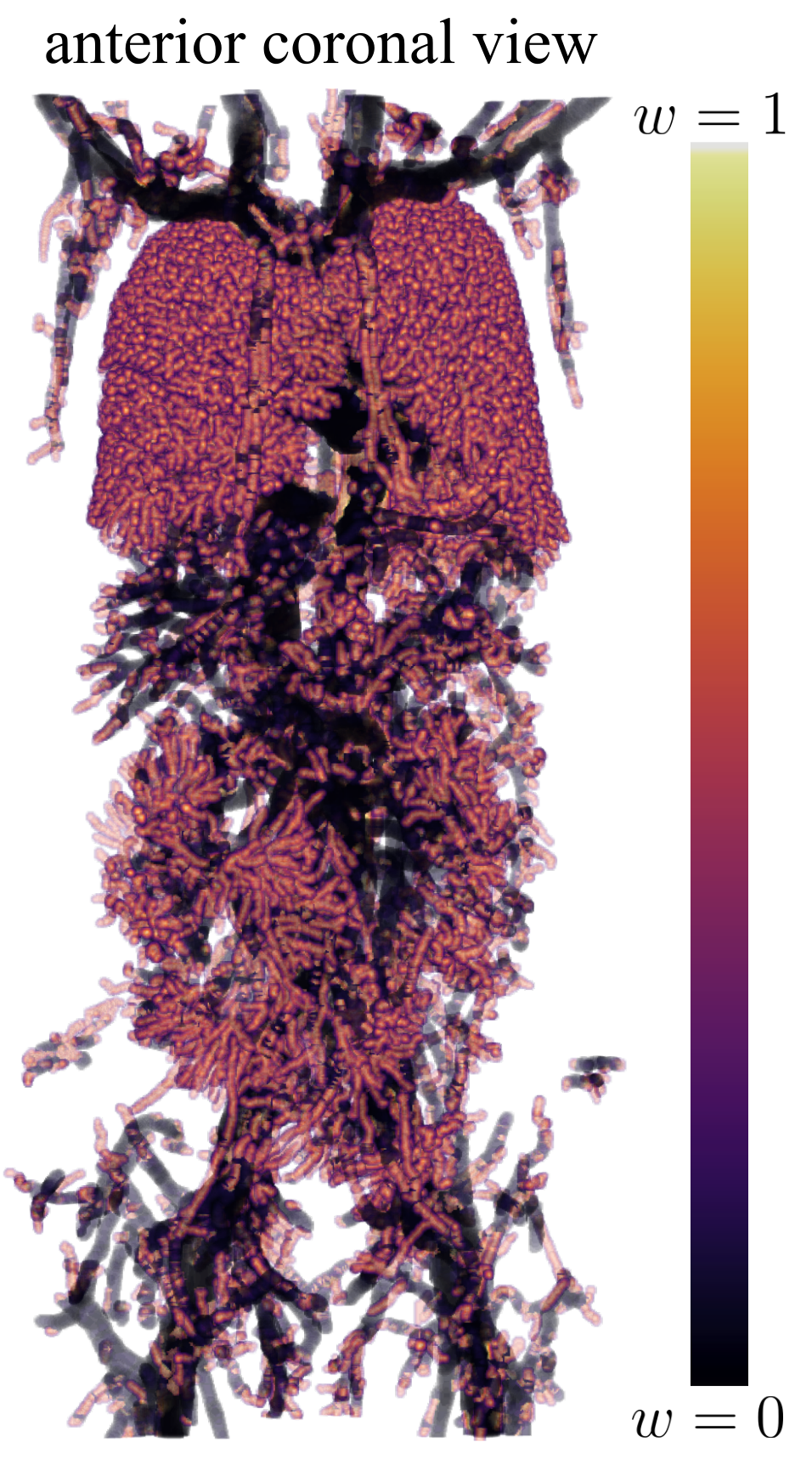}
\vspace{-1em}
\caption{
Unscaled weighting map $w$ derived from a training sample.
%($d_\text{FPP} = 3$; L15 in Algorithm~\ref{alg:weights} for reference). Our weighting scheme assigns higher weights to minuscule vessels, while larger vessels receive comparatively lower weights.
}
\vspace{-1em}
\label{fig:w_real}
\end{wrapfigure}

Our voxel-level weights $w$ are derived from the ground truth segmentation mask $y$ and conceptually described in Algorithm~\ref{alg:weights} and Fig.~\ref{fig:w}.
First, we extract vessel centerlines $c$ by skeletonizing the segmentation mask $y$. We then estimate for each centerline segment the distance to the vessel wall, resulting in an accurate radius estimate $r_c$ for the complete centerline.
Next, these radius estimates are propagated from the centerline to all foreground voxels by assigning each voxel the radius of its nearest centerline segment. The radius values $r$ are subsequently clipped and converted into local vessel cross-sectional areas $a$ by counting the number of voxels whose centers lie within a circle of the corresponding radius.
To compensate for the over-representation of large vessels, we define the voxel-wise weights as the inverse of the estimated vessel cross-sectional area, thereby ensuring that thinner vessels receive higher weights (see Fig.~\ref{fig:w_real}). This yields a radius-aware weighting scheme that balances contributions across vessel scales.

To further enforce accurate vessel borders and mitigate the challenge of over-segmentation resulting from over-weighted foreground voxels, we introduce an FP penalty (FPP) applied directly at the vessel boundary that propagates previously estimated weights into the surrounding background. Specifically, we apply a linear decay within a predefined distance $d_\text{FPP}$. This encourages the model to penalize potential FP predictions near vessel boundaries more strongly.
Finally, the weight map is scaled and shifted using a global scale factor $\alpha_w$ and offset, yielding the final voxel-level weight map $w$.

In summary, \lossname{} provides a natural and computationally efficient solution, avoiding the over-simplifying assumption of, \eg, reducing vasculature to centerlines, while incorporating relevant vessel-specific biases into the commonly utilized DiceCE compound loss. \lossname{} is further robust to missing annotations and well suited for settings where vessels are difficult to annotate, as unlabeled vessels treated as background are not strongly penalized due to the low background weighting (see Fig.~\ref{fig:w}, omitted vessel). We intentionally design \lossname{} to rely solely on a few parameters, minimizing the need for parameter tuning while allowing flexible control. One can, \eg, steer the aggressiveness of \lossname{} via $\alpha_w$ (see~\ref{app:alpha_w}), with $\alpha_w =0 $ reducing the loss to a default DiceCE implementation, and adjust the coverage of regions surrounding vessels via $d_\text{FPP}$. 

We further build upon \lossname{}'s intuition to provide the matching \metricname{} metric, accurately measuring the performance of blood vessel segmentation algorithms by accounting for intrinsic class imbalance while ensuring sharp vessel borders.
%In contrast to the commonly used, vessel-specific clDice metric, \metricname{} relies on our above described vessel-specific prior, ranging beyond reducing vasculature to centerlines. 
To construct \metricname{}, we adjust the Dice metric with our unclipped, unscaled, and unshifted voxel-level weights 
%(see Algorithm~\ref{alg:weights}, L7) 
as $\text{\metricname{}} = \frac{2 \sum_{i} w_i \hat{y}_i y_i}{\sum_{i} w_i (\hat{y}_i + y_i)}$.
%as $\text{\metricname{}} = \left( 2 \sum_{i} w_i \hat{y}_i y_i \right) / \left( \sum_{i} w_i (\hat{y}_i + y_i) \right)$.

\section{Implementation Details}\label{sec:impl}
We train \methodname{} for 312.500 iterations on a single NVIDIA RTX 5000 Ada GPU (32 GB VRAM) optimizing \lossname{} with $\lambda_\text{Dice}$ and $\lambda_\text{CE}$ set to 0.5, $d_\text{FPP}$ set to 4, $\alpha_w$ set to 100, and $r_\text{cutoff}$ set to 5. Omitting $r_\text{cutoff}$, \metricname{} is evaluated with $d_\text{FPP}$ set to 4, and $\alpha_w$ set to 1.
VesselFM originally relies on the relatively light-weight DynUNet architecture~\cite{wittmann2025vesselfm,cardoso2022monai}  (31.4 M param.). Given that the task of segmenting the complete vascular network is extremely complex, we exchange vesselFM's architecture for fine-tuning to match the higher-capacity nnU-Net ResEnc config~\cite{isensee2024nnu} (102.3 M param.) and retrain it using its original data sources~\cite{wittmann2025vesselfm} (see Table~\ref{tab:quant_res}, \textit{arch.}). During training, $p_\text{syn}$ is set to 20 \% in iteration 1. Given that, the fine-tuned \methodname{} demonstrates strong performance after already the first iteration, we omit TTA in Step 1) due to the drastic increase in inference time, omit Step 2) completely, adjust post-processing parameter in Step 3), and set $p_\text{syn}$ to zero in Step 5). Additional implementation details are mentioned in~\ref{app:impl}.

\begin{table}[t]
\centering
\small
\caption{
Quantitative Experiments. We compare \methodname{} to baseline methods (\textit{baselines}), evaluate loss functions (\textit{loss function}), ablate \lossname{}'s components (\textit{loss abl.}), experiment with nnUNet's updated architecture (\textit{arch.}), and analyze performance across training iterations (\textit{itera.}).
}
\label{tab:quant_res}
\begin{tabular}{l l c c c c c}
\toprule
& Method (Config) / Ablations & $\text{TubeDice}\uparrow$ & $\text{TPR}\uparrow$ & $\text{Dice}\uparrow$ & $\text{IoU}\uparrow$ & $\text{clDice}\uparrow$ \\

\midrule
\multirow{5}{*}{\rotatebox[origin=c]{90}{\textit{loss function}}}
& \methodname{} (TubeLoss) & \textbf{88.85}{\tiny$\pm$0.12} & \textbf{93.33}{\tiny$\pm$0.81} & \textbf{90.44}{\tiny$\pm$1.03} & \textbf{83.53}{\tiny$\pm$0.42} & 86.43{\tiny$\pm$0.83} \\
& \methodname{} (DiceCE) & \underline{84.44}{\tiny$\pm$1.01} & 87.95{\tiny$\pm$1.55} & \underline{90.33}{\tiny$\pm$0.93} & \underline{82.37}{\tiny$\pm$1.53} & \underline{87.82}{\tiny$\pm$0.46} \\
& \methodname{} (soft-clDice~\cite{shit2021cldice}) & 76.13{\tiny$\pm$1.32} & 86.96{\tiny$\pm$1.56} & 84.63{\tiny$\pm$1.71} & 73.38{\tiny$\pm$2.54} & \textbf{90.69}{\tiny$\pm$0.39} \\
& \methodname{} (cbDice~\cite{shi2024centerline}) & 78.23{\tiny$\pm$2.20} & 83.43{\tiny$\pm$2.49} & 86.71{\tiny$\pm$1.67} & 76.56{\tiny$\pm$2.59} & 82.95{\tiny$\pm$1.89} \\
& \methodname{} (Skel. Recall~\cite{kirchhoff2024skeleton}) & 75.37{\tiny$\pm$0.51} & \underline{92.01}{\tiny$\pm$0.45} & 79.18{\tiny$\pm$1.26} & 65.54{\tiny$\pm$1.74} & 79.18{\tiny$\pm$1.27} \\

\midrule
\multirow{4}{*}{\rotatebox[origin=c]{90}{\textit{loss abl.}}}
& $r_\text{cutoff}$ \ding{51}\quad  FPP \ding{51}\quad  $1/a$ \ding{51} & \textbf{88.85}{\tiny$\pm$0.12} & 93.33{\tiny$\pm$0.81} & \textbf{90.44}{\tiny$\pm$1.03} & \textbf{83.53}{\tiny$\pm$0.42} & \underline{86.43}{\tiny$\pm$0.83} \\
& $r_\text{cutoff}$ \ding{55}\hspace{2.7mm} FPP \ding{51}\quad  $1/a$ \ding{51} & \underline{87.89}{\tiny$\pm$1.27} & 91.33{\tiny$\pm$3.04} & \underline{90.43}{\tiny$\pm$0.97} & \underline{82.53}{\tiny$\pm$1.61} & \textbf{86.61{\tiny$\pm$0.67}} \\
& $r_\text{cutoff}$ \ding{55}\hspace{2.7mm}  FPP \ding{55}\hspace{2.9mm}  $1/a$ \ding{51} & 65.24{\tiny$\pm$8.64} & \underline{96.62}{\tiny$\pm$0.51} & 72.45{\tiny$\pm$5.81} & 57.02{\tiny$\pm$7.01} & 78.32{\tiny$\pm$0.41} \\
& $r_\text{cutoff}$ \ding{55}\hspace{2.7mm}  FPP \ding{55}\hspace{2.9mm}  $1/a$ \ding{55} & 61.43{\tiny$\pm$6.89} & \textbf{98.37}{\tiny$\pm$0.31} & 65.45{\tiny$\pm$4.30} & 48.74{\tiny$\pm$4.73} & 73.65{\tiny$\pm$0.63} \\

\midrule
\multirow{2}{*}{\rotatebox[origin=c]{90}{\textit{arch.}}}
& \methodname{} (ResEnc~\cite{isensee2024nnu}) & \textbf{88.85}{\tiny$\pm$0.12} & \textbf{93.33}{\tiny$\pm$0.81} & \textbf{90.44}{\tiny$\pm$1.03} & \textbf{83.53}{\tiny$\pm$0.42} & \textbf{86.43}{\tiny$\pm$0.83} \\
& \methodname{} (DynUNet~\cite{cardoso2022monai}) & \underline{87.48}{\tiny$\pm$0.29} & \underline{92.63}{\tiny$\pm$0.78} & \underline{88.80}{\tiny$\pm$0.89} & \underline{80.68}{\tiny$\pm$0.47} & \underline{85.48}{\tiny$\pm$1.00} \\

\midrule
\multirow{3}{*}{\rotatebox[origin=c]{90}{\textit{itera.}}}
& \methodname{} (iteration 1) & 88.85{\tiny$\pm$0.12} & 93.33{\tiny$\pm$0.81} & 90.44{\tiny$\pm$1.03} & 83.53{\tiny$\pm$0.42} & 86.43{\tiny$\pm$0.83} \\
& \methodname{} (iteration 2) & \underline{93.53}{\tiny$\pm$1.34} & \textbf{96.28}{\tiny$\pm$0.34} & \underline{93.99}{\tiny$\pm$0.19} & \underline{88.77}{\tiny$\pm$0.40} & \underline{90.15}{\tiny$\pm$0.59} \\
& \methodname{} (iteration 3) & \textbf{93.71}{\tiny$\pm$0.47} & \underline{95.87}{\tiny$\pm$0.49} & \textbf{94.32}{\tiny$\pm$0.12} & \textbf{89.24}{\tiny$\pm$0.21} & \textbf{90.45}{\tiny$\pm$0.64} \\

\midrule
\multirow{5}{*}{\rotatebox[origin=c]{90}{\textit{baselines}}}
& \methodname{} (iteration 3) & \textbf{93.71}{\tiny$\pm$0.47} & \textbf{95.87}{\tiny$\pm$0.49} & \textbf{94.32}{\tiny$\pm$0.12} & \textbf{89.24}{\tiny$\pm$0.21} & \textbf{90.45}{\tiny$\pm$0.64} \\
& CADS-model~\cite{xu2025cads} & 20.89{\tiny$\pm$6.20} & 32.66{\tiny$\pm$0.75} & 48.77{\tiny$\pm$0.74} & 32.25{\tiny$\pm$0.65} & 10.77{\tiny$\pm$0.52} \\
& TotalSegmentator~\cite{wasserthal2023totalsegmentator,hinck2025automatic} & \underline{59.29}{\tiny$\pm$0.59} & \underline{48.80}{\tiny$\pm$0.96} & \underline{63.77}{\tiny$\pm$0.76} & \underline{46.81}{\tiny$\pm$0.82} & \underline{62.11}{\tiny$\pm$3.86} \\
& VoxTell~\cite{rokuss2025voxtell} & 22.37{\tiny$\pm$3.32} & 36.66{\tiny$\pm$0.42} & 53.06{\tiny$\pm$0.52} & 36.11{\tiny$\pm$0.48} & 17.48{\tiny$\pm$0.34} \\
& BiomedParse (V2)~\cite{zhao2025foundation,zhao2025boltzmann} & 19.55{\tiny$\pm$4.93} & 26.75{\tiny$\pm$1.09} & 22.89{\tiny$\pm$0.94} & 12.92{\tiny$\pm$0.60} & 10.38{\tiny$\pm$0.55} \\
\bottomrule
\end{tabular}
\end{table}

\section{Evaluation of \methodname{} and \lossname{}}
We systematically evaluate \methodname{} and \lossname{} by comparing to alternative loss functions, ablating components of \lossname{}, analyzing performance across training iterations and architectures, and benchmarking against baseline methods. Results obtained on the test set are reported in Table~\ref{tab:quant_res}. Unless otherwise specified, we conduct experiments on the first training iteration, as its results guided our overall design decisions. We further exclude synthetic data in first-iteration experiments to isolate performance on the blood vessel segmentation task, avoiding potential confounding effects.

\paragraph{Loss Function:} We compare our introduced \lossname{} to the standard DiceCE loss, soft-clDice, cbDice, and the Skeleton Recall loss (see Table~\ref{tab:quant_res}, \textit{loss function}). All baseline loss functions were configured according to their recommendations. Notably, these configurations consist of compound losses that combine multiple objectives. \lossname{} achieves best \metricname{}, TPR, Dice, and IoU scores outperforming all baselines by a large margin, indicating its effectiveness for cardiovascular segmentation. We hypothesize that clDice scores appear artificially low, given that the model trained on \lossname{} segments tiny free-floating blood vessels, which are omitted from the ground truth via small, unconnected component removal (see Fig.~\ref{fig:training}, Step 3).
We further demonstrate \lossname{}'s superior performance qualitatively in~\ref{app:qual_res_loss}. We additionally ablate main components of \lossname{}. To this end, we omit radius clipping, omit our introduced FPP, and finally omit inverse-area weighting, assigning $w = 1$ instead before scaling and shifting (see Table~\ref{tab:quant_res}, \textit{loss abl.}). One can observe that all components meaningfully contribute to \lossname{}'s performance, validating our rationale.

\begin{figure*}[t!]
\centering
\includegraphics[width=0.85\linewidth]{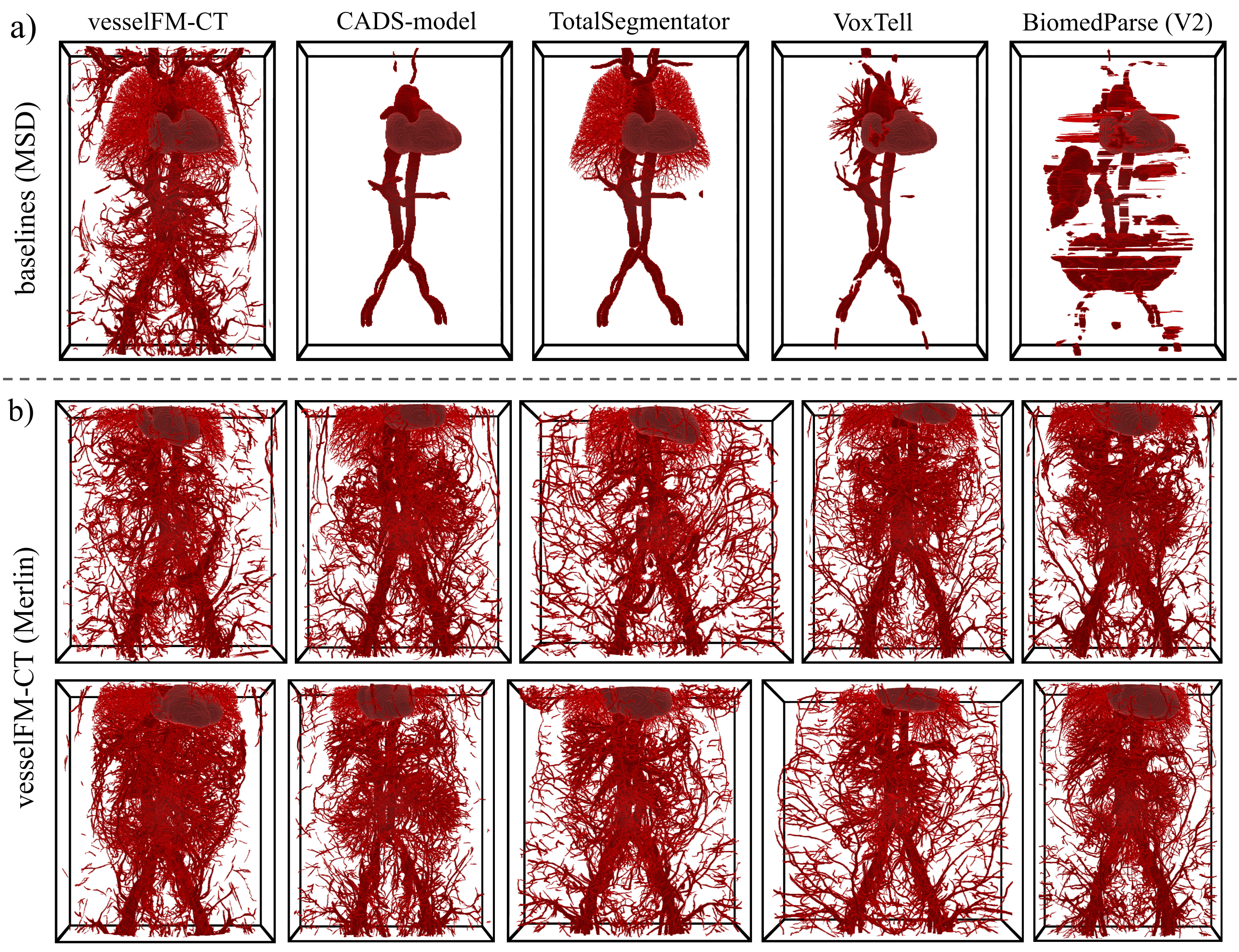}
\vspace{-0.5em}
\caption{
a) Qualitative comparison of \methodname{} with baseline methods on a sample from our test set.
b) Exemplary results of \methodname{} on the Merlin dataset (better viewed zoomed in).
}
\label{fig:qual_res}
\end{figure*} 

\paragraph{Architecture and Training Iterations:}
We compare the recently proposed nnU-Net ResEnc config~\cite{isensee2024nnu} to base vesselFM's light-weight DynUNet architecture~\cite{wittmann2025vesselfm,cardoso2022monai}, and find an overall performance increase (see Table~\ref{tab:quant_res}, \textit{arch.}), indicating its increased capacity. \Methodname{} is trained in an iterative process consisting of three iterations. We observe, as expected, an improvement over training iterations, which gradually diminishes toward the final iteration (see Table~\ref{tab:quant_res}, \textit{iterations}).

\paragraph{Baseline Models:}
We compare \methodname{} to anatomical models capable of segmenting a broad range of cardiovascular structures. Specifically, we evaluate two general-purpose CT segmentation models (CADS-model~\cite{xu2025cads} and TotalSegmentator~\cite{wasserthal2023totalsegmentator,hinck2025automatic} with its vascular subtasks~\cite{poletti2022automated,simpson2019large,walter2024segmentation}) and two free-text universal medical segmentation models (VoxTell~\cite{rokuss2025voxtell} and BiomedParse~\cite{zhao2025foundation,zhao2025boltzmann}). We aggregate vascular labels from the general-purpose CT segmentation models and outputs of 21 vessel-specific prompts from the free-text models to obtain final predictions (see~\ref{app:impl} for details). Quantitative results are reported in Table~\ref{tab:quant_res} (\textit{baselines}), while qualitative results are shown in Fig.~\ref{fig:qual_res}a. \Methodname{} outperforms all baselines by a wide margin, demonstrating its expressiveness.

\section{Downstream Experiments}\label{sec:down}
\begin{wrapfigure}{r}{0.32\textwidth}
% \vspace{-1em}
\centering
\includegraphics[width=\linewidth]{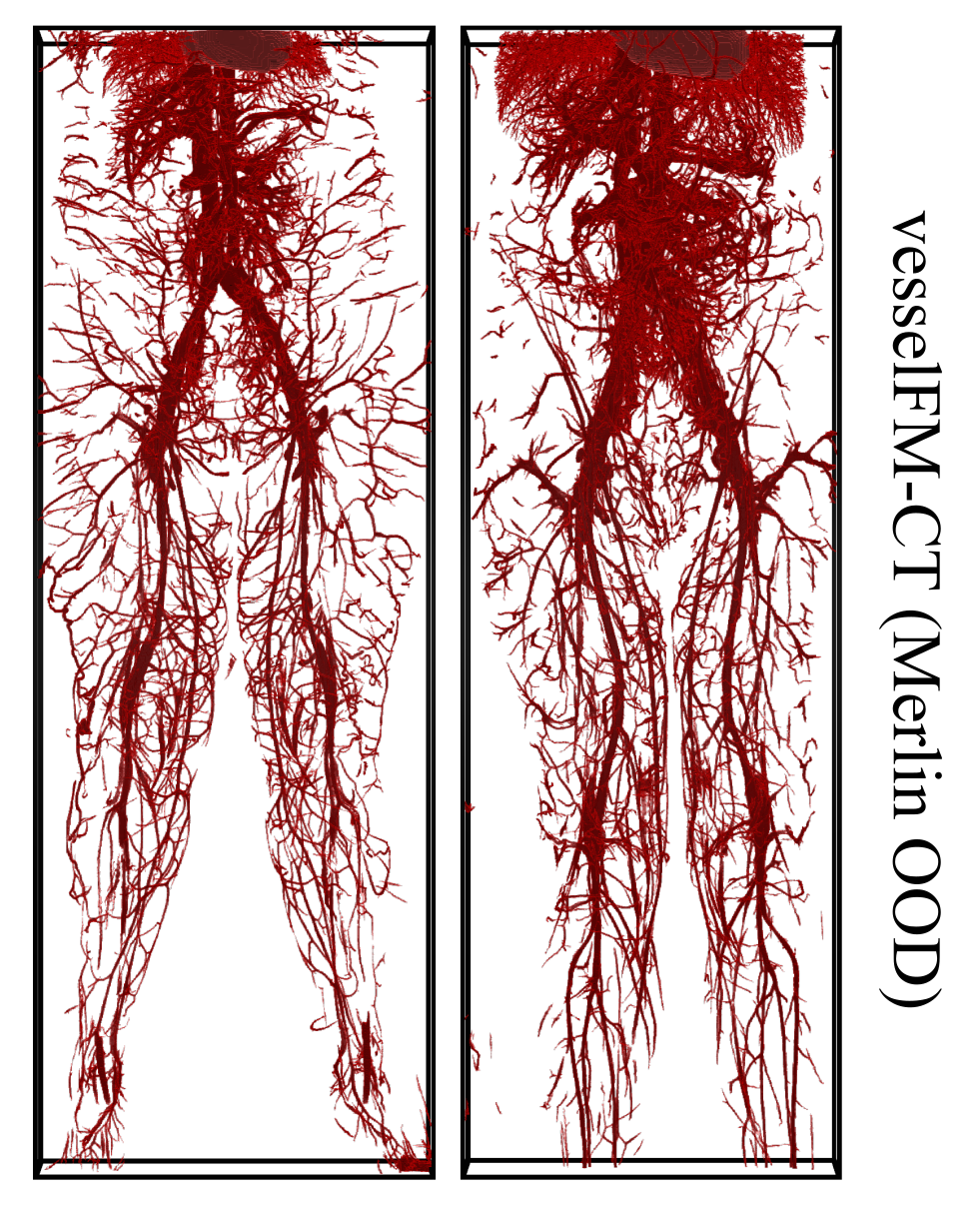}
\vspace{-1em}
\caption{
Results on OOD\\ anatomical structures.
}
\vspace{-0.5em}
\label{fig:qual_res_ood}
\end{wrapfigure}
We assess \methodname{}'s clinical and technical perspectives via two downstream experiments: disease (or condition) classification and synthetic CT image generation. Both experiments are conducted on the Merlin dataset~\cite{blankemeier2026merlin}, which consists of abdominal CT scans, matching radiology reports, and relevant metadata. Derived from the reports, Merlin further provides disease labels (-1 missing, 0 negative, and 1 positive). We subsample Merlin to solely contain scans of appropriate slice thickness (1 or 1.25 mm), portal venous contrast phase, and explicit presence or absence of diseases that have or may have an impact on the cardiovascular system (\eg, thrombosis or abdominal aortic aneurysm). Finally, we rebalanced the subset to mitigate class imbalance. As a result, the final subset comprises $\sim$3,000 CT images with corresponding disease labels.
Applying \methodname{}, trained on MSD, to CT images from Merlin highlights its robustness, as it segments cardiovascular structures in unprecedented detail (see Fig.~\ref{fig:qual_res}b and~\ref{app:more_qual_res}). \Methodname{} is not only robust to new data sources but also generalizes exceptionally well to out-of-distribution (OOD) anatomies not seen during training, such as the lower extremities (see Fig.~\ref{fig:qual_res_ood}).

\begin{figure*}[t]
\centering
\includegraphics[width=\linewidth]{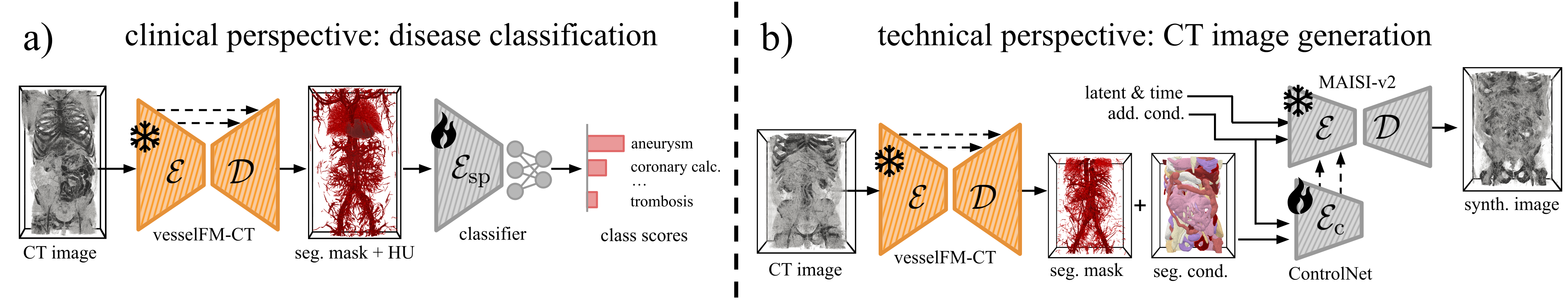}
\vspace{-1em}
\caption{
Experimental setup for downstream tasks. We experiment with disease classification from CT images and enrich existing, conditional CT image generation methods with our vasculature.
}
\label{fig:downstream}
\end{figure*}

\paragraph{Disease Classification:} We showcase clinical perspectives of \methodname{} by predicting selected diseases from the Merlin dataset based on vascular structure. To this end, we apply \methodname{} to CT images and feed the predicted masks together with their overlaid HU values into a sparse encoder, followed by a final MLP (see Fig.~\ref{fig:downstream}a). The sparse encoder uses sparse convolutional layers~\cite{prabhakar2026sparse} for increased efficiency and performance. We compare our approach to the same architecture, swapping the sparse convolutional layers with dense convolutional layers, operating on the CT image instead.
We find that the classifier trained on \methodname{}'s predicted segmentation masks performs superior on cardiovascular diseases compared to the classifier trained directly on the CT images (average over cardiovascular diseases: \textbf{80.71} vs. 75.70 AUC). The \methodname{}-based classifier shows great improvements in detecting abdominal aortic aneurysms (\textbf{90.47} vs. 79.11 AUC) and thrombosis (\textbf{58.78} vs. 50.95 AUC), both of which are clinically challenging to identify from intensity information alone due to their reliance on subtle structural and contextual cues. This suggests that incorporating segmentation-derived structural priors improves the classifier's ability to capture vascular morphology and abnormal vessel geometry. We further observe a performance increase on coronary calcification (\textbf{76.38} vs. 68.97 AUC) and aortic valve calcification (\textbf{92.86} vs. 89.43 AUC). The image-based classifier performs superior on atherosclerosis (\textbf{90.06} vs. 85.08 AUC), a disease which is highly visible in CT images due to strong and heterogeneous intensity patterns, and therefore may not benefit from segmentation-based abstraction. 
%Interestingly, we find that the \methodname{}-based classifier performs superior in detecting metastasic disease (\textbf{77.61} vs. 73.15 AUC).
Additional details are mentioned in~\ref{app:down}.

% \begin{wraptable}{r}{0.5\textwidth}
% \centering
% \small
% \vspace{-1em}
% \caption{Disease classification on our curated Merlin subset. We report AUC values on the test set.}
% \vspace{0.5em}
% \label{tab:quant_res_down}
% \begin{tabular}{ll c c}
% \toprule
% & Disease / Condition & Ours & Base. \\

% \midrule
% \multirow{5}{*}{\rotatebox[origin=c]{90}{\textit{cardiovasc.}}}
% &Atherosclerosis                    & 85.08 & \textbf{90.06} \\
% &Coronary Calc.                    & \textbf{76.38} & 68.97 \\
% &Aortic Valve Calc.                & \textbf{92.86} & 89.43 \\
% &Abd. Aortic Aneurysm              & \textbf{90.47} & 79.11 \\
% &Thrombosis                        & \textbf{58.78} & 50.95 \\

% \midrule
% \multirow{4}{*}{\rotatebox[origin=c]{90}{\textit{other}}}
% &Metastatic Disease                & \textbf{77.61} & 73.15 \\
% &Renal Hypodensities               & \textbf{67.12} & 64.56 \\
% &Pleural Effusion                  & \textbf{93.87} & 87.51 \\
% &Splenomegaly                      & \textbf{79.89} & 72.26 \\
% \bottomrule
% \end{tabular}
% \end{wraptable}

\paragraph{CT Generation:} 
\begin{wrapfigure}{r}{0.4\textwidth}
\vspace{-1em}
\centering
\includegraphics[width=\linewidth]{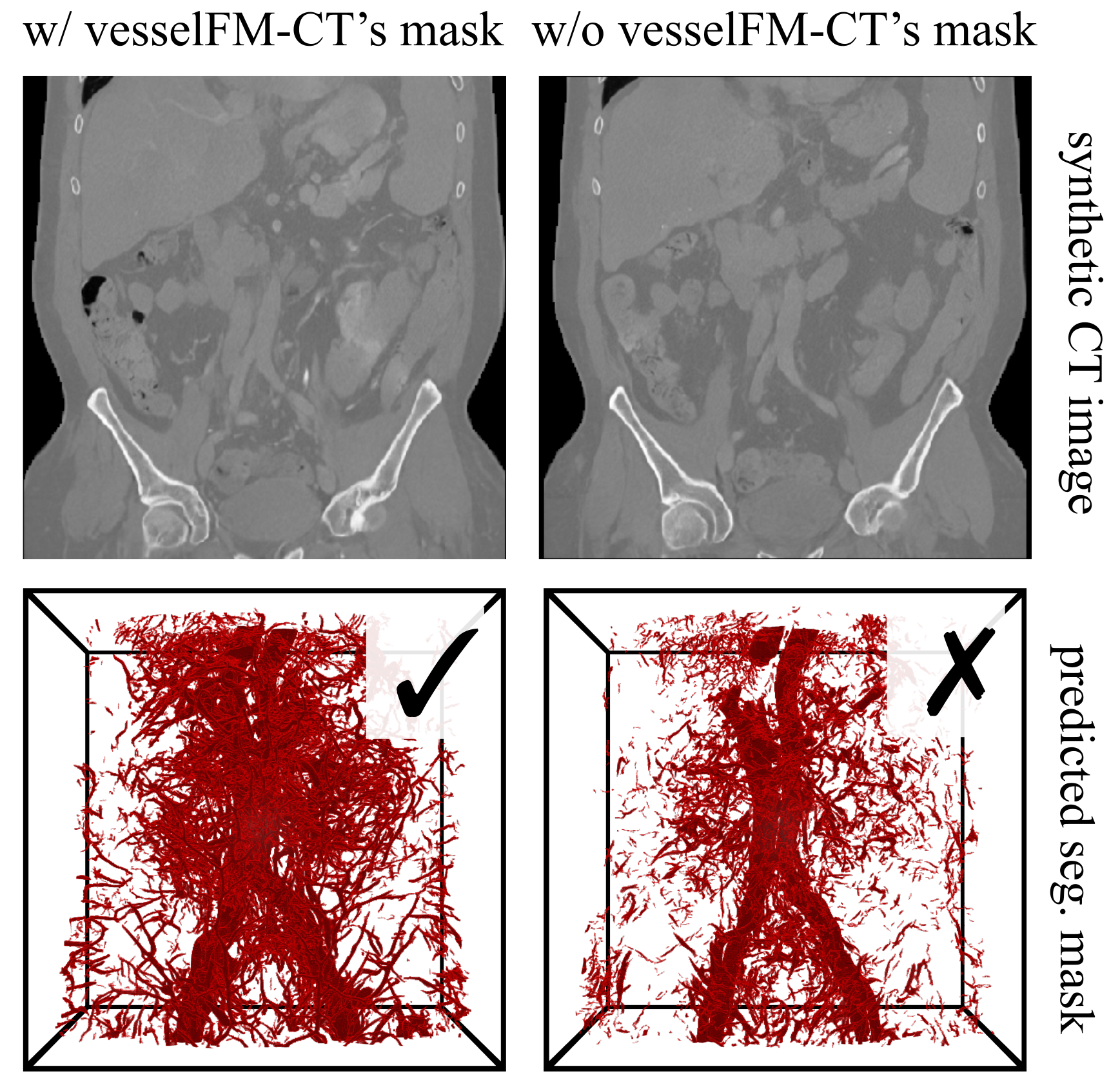}
\vspace{-1em}
\caption{
MAISI-v2 conditioned with (left) and without (right) \methodname{}'s segmentation masks. 
}
\vspace{-0.5em}
\label{fig:qual_res_maisi}
\end{wrapfigure}
We further showcase technical perspectives of \methodname{} by refining generation capabilities of the state-of-the-art CT image generation model MAISI-v2~\cite{zhao2025boltzmann,guo2025maisi} (see Fig.~\ref{fig:downstream}b). MAISI-v2 is a rectified flow-based latent diffusion model relying on ControlNet~\cite{zhang2023adding} conditioning to incorporate anatomical structures during generation. 
However, while most structures are used during conditioning, the cardiovascular system in its entirety is not included. We therefore add \methodname{}'s predicted segmentation masks on top of MAISI-v2's ControlNet conditioning, aiming at generating synthetic images that faithfully capture the previously missing cardiovascular network. We conduct two experiments fine-tuning the ControlNet with (see Fig.~\ref{fig:qual_res_maisi}, left) and without \methodname{}’s segmentation masks (see Fig.~\ref{fig:qual_res_maisi}, right ) derived from our Merlin subset. We find that the version additionally conditioned on our vascular segmentation mask improves performance, generating more anatomically accurate and spatially coherent vascular structures (see Fig.~\ref{fig:qual_res_maisi}, first row). To further support this claim, we apply \methodname{} to the generated images (see Fig.~\ref{fig:qual_res_maisi}, second row), aiming at assessing the plausibility of the cardiovascular system in the synthetic images. We find that the synthetic image generated by the MAISI-v2 variant conditioned without our masks completely fails to capture the vascular structure. Additional details are mentioned in~\ref{app:down}.

\section{Conclusion and Outlook}
In this work, we present \methodname{}, the first model capable of comprehensive segmentation of the cardiovascular network in unprecedented detail. \Methodname{}'s capabilities stem from a complex, multi-step training process optimizing our proposed \lossname{} loss function, scaling blood vessel segmentation to the complete cardiovascular system. We demonstrate state-of-the-art results on a clinically relevant task, thoroughly evaluate \methodname{}'s components, demonstrate robustness to unseen data sources and even anatomical structures, and show in our experiments that \methodname{} unlocks both clinical and technical perspectives. We further anticipate that \methodname{} will advance systemic analysis of the cardiovascular network and impact domains where precise knowledge of vascular anatomy is critical to minimize risk and optimize patient outcomes (\eg, tumor growth modeling, hemodynamic simulation, radiation therapy planning, or surgical planning).

\begin{ack}
This work was supported by the Helmut Horten Foundation.
SS and BM acknowledge support from the Swiss National Science Foundation via grant 232553.
The authors declare no competing interests.
\end{ack} 

% \begin{ack}
% Use unnumbered first level headings for the acknowledgments. All acknowledgments
% Do {\bf not} include this section in the anonymized submission, only in the final paper. You can use the \texttt{ack} environment provided in the style file to automatically hide this section in the anonymized submission.
% \end{ack}

\bibliographystyle{plain}
{
\small
\bibliography{references}
}

\newpage
\appendix
\section{Technical Appendices and Supplementary Material}

\subsection{\lossname{} Implementation Details}\label{app:loss}
\lossname{} is defined as:

\begin{equation}
\mathcal{L}_{\text{Tube}} = \lambda_{\text{Dice}} \, \mathcal{L}_{\text{Dice}}^{(w)} + \lambda_{\text{CE}} \, \mathcal{L}_{\text{CE}}^{(w)},
\end{equation}

where $\mathcal{L}_{\text{Dice}}^{(w)}$ represents the weighted Dice loss, $\mathcal{L}_{\text{CE}}^{(w)}$ the weighted binary CE loss, and $\lambda$ values describe weighting coefficients. Following MONAI's~\cite{cardoso2022monai} implementation, we augment the Dice loss with weights as:

\begin{equation}
\mathcal{L}_{\text{Dice}}^{(w)} = 1 - \frac{2 \, \mathrm{TP}^{(w)}}{2\mathrm{TP}^{(w)} + \mathrm{FP}^{(w)} + \mathrm{FN}^{(w)}},
\end{equation}

where true positives (TP), false positives (FP), and false negatives (FN) are defined as:

\begin{equation}
\mathrm{TP}^{(w)} = \frac{\text{Pred}^{(w)} + \text{GT}^{(w)} - \text{Diff}^{(w)}}{2},\quad
\mathrm{FP}^{(w)} = \text{Pred}^{(w)} - \mathrm{TP}^{(w)},\quad
\mathrm{FN}^{(w)} = \text{GT}^{(w)} - \mathrm{TP}^{(w)},
\end{equation}

with predictions (Pred), ground truth (GT), and their difference (Diff) being weighted according to:

\begin{equation}
\text{Pred}^{(w)} = \sum_{i=1}^{N} w_i \hat{y}_i,\quad
\text{GT}^{(w)} = \sum_{i=1}^{N} w_i y_i,\quad
\text{Diff}^{(w)} = \sum_{i=1}^{N} w_i (\hat{y}_i - y_i).
\end{equation}

Similarly, we employ the standard, weighted binary CE loss, formulated as:

\begin{equation}
\mathcal{L}_{\text{CE}}^{(w)} = - \sum_{i=1}^{N} w_i \Big[ y_i \, \log(\hat{y}_i) + (1 - y_i) \, \log(1 - \hat{y}_i) \Big].
\end{equation}

\subsection{Effect of Synthetic Data}\label{app:syn}
\begin{figure}[t]
\centering
\includegraphics[width=0.7\linewidth]{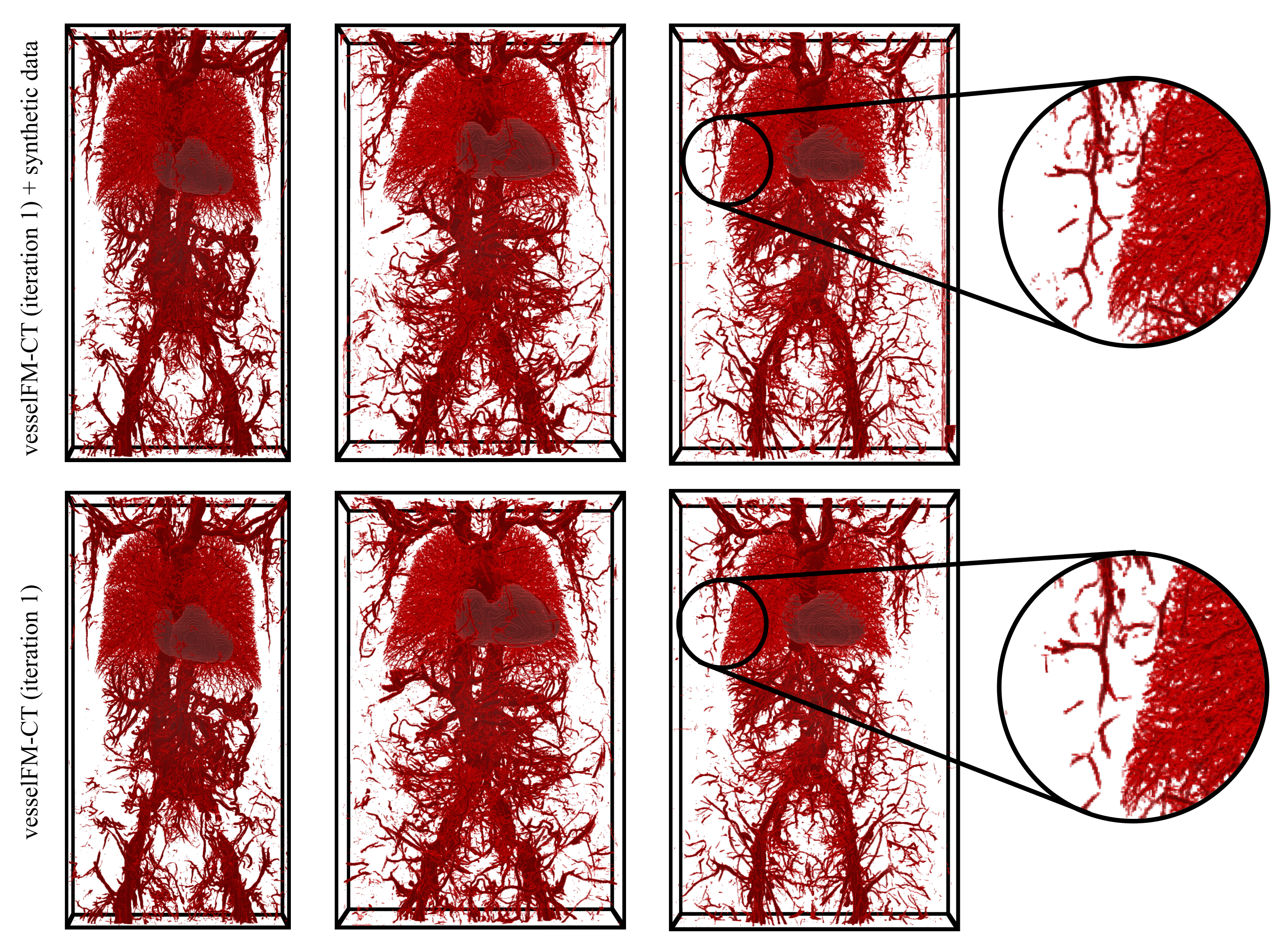}
\vspace{-0.5em}
\caption{
\Methodname{} trained with (first row) and without (second row) synthetic data. Given that synthetic data is utilized solely in the first training iteration, we compare models from the first iteration. Involving synthetic data during training increases sensitivity towards tubular structures.
}
\label{fig:app_syn}
\end{figure}

We use synthetic data in the initial training iteration to boost \methodname{}'s sensitivity towards tubular structures.
Although this increases the number of FPs, it accelerates the labor-intensive manual annotation process, as removing FPs is considerably easier and faster than correcting FNs. Consequently, the use of synthetic data enables efficient generation of high-fidelity label maps. We demonstrate the effect of synthetic data in the initial training iteration by comparing predictions from models trained with (first row) and without (second row) the synthesized vasculature in Fig.~\ref{fig:app_syn}.

\subsection{\texorpdfstring{Effect of Adjusting $\alpha_w$ in \lossname{}}:}\label{app:alpha_w}
\begin{figure}[h]
\centering
\includegraphics[width=0.6\linewidth]{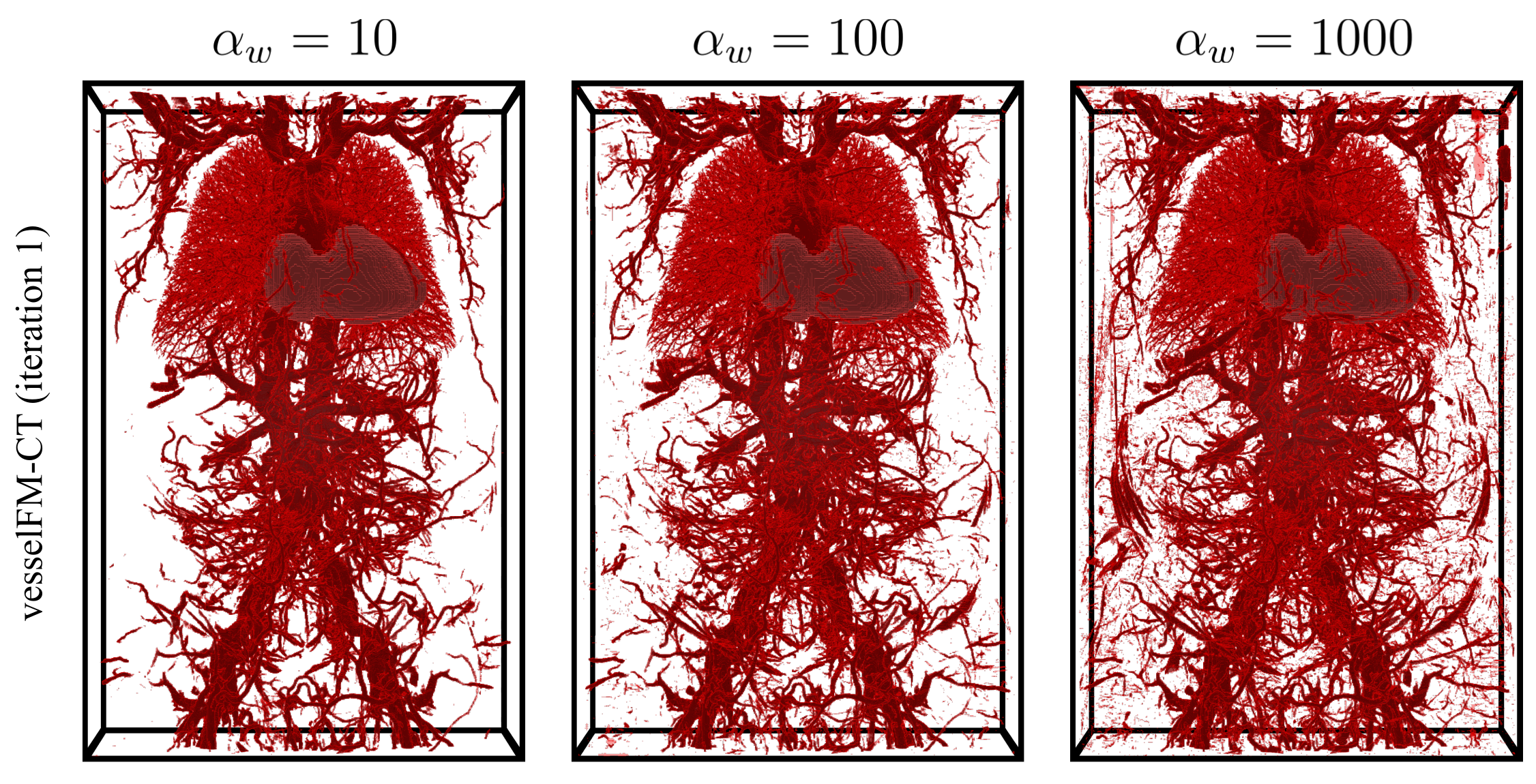}
\vspace{-0.5em}
\caption{
\Methodname{} trained with different configurations of \lossname{}. We qualitatively compare \lossname{}'s scale factor $\alpha_w = 100$, to $\alpha_w = 10$, and $\alpha_w = 1000$.
}
\label{fig:app_alphaw}
\end{figure}

Adjusting $\alpha_w$ in \lossname{} can steer its aggressiveness towards small blood vessels. Lower $\alpha_w$ values reduce the influence of the weighting scheme ($\alpha_w = 0$ renders \lossname{} equivalent to the standard DiceCE loss), whereas excessively high $\alpha_w$ values can lead to overemphasis on tiny structures, resulting in a drastic increase in FPs and therefore noisy predictions.
\begin{table}[h]
\centering
\small
\caption{
Ablation on the effect of \lossname{}' scale factor $\alpha_{w}$.
}
\label{tab:quant_res_alpha}
\begin{tabular}{l c c c c c}
\toprule
Method (Config) & $\text{TubeDice}\uparrow$ & $\text{TPR}\uparrow$ & $\text{Dice}\uparrow$ & $\text{IoU}\uparrow$ & $\text{clDice}\uparrow$ \\
\midrule
\methodname{} ($\alpha_{w} = 10$) & 86.29{\tiny$\pm$1.19} & 90.63{\tiny$\pm$2.29} & \textbf{90.50}{\tiny$\pm$1.13} & \underline{82.66}{\tiny$\pm$1.89} & \textbf{87.53}{\tiny$\pm$0.18} \\
\methodname{} ($\alpha_{w} = 100$) & \textbf{88.85}{\tiny$\pm$0.12} & \textbf{93.33}{\tiny$\pm$0.81} & \underline{90.44}{\tiny$\pm$1.03} & \textbf{83.53}{\tiny$\pm$0.42} & \underline{86.43}{\tiny$\pm$0.83} \\
\methodname{} ($\alpha_{w} = 1000$) & \underline{88.54}{\tiny$\pm$1.22} & \underline{92.78}{\tiny$\pm$2.64} & 88.59{\tiny$\pm$0.48} & 79.52{\tiny$\pm$0.76} & 81.07{\tiny$\pm$3.25} \\
\bottomrule
\end{tabular}
\end{table}
To demonstrate the effect of $\alpha_w$, we experiment with three values: $10$, $100$, and $1000$. Quantitative results are shown in Table~\ref{tab:quant_res_alpha}, while qualitative results are presented in Fig.~\ref{fig:app_alphaw}. We quantitatively and qualitatively find an $\alpha_w$ value of $100$ ideal for the task of cardiovascular network segmentation. Consequently, we set $\alpha_w = 100$ in our experiments.

\begin{figure}[h]
\centering
\includegraphics[width=\linewidth]{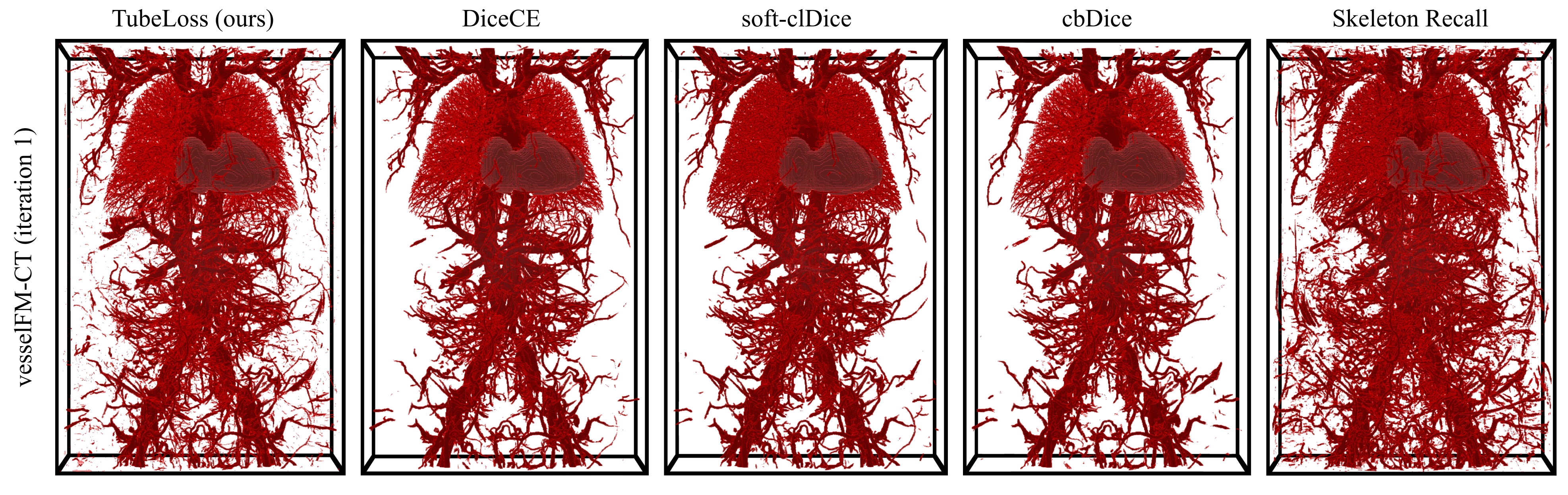}
\vspace{-1em}
\caption{
Comparison of \methodname{} trained with different loss functions.
}
\label{fig:app_loss}
\end{figure}

\subsection{Qualitative Comparison of Loss Functions:}\label{app:qual_res_loss}
To complement quantitative results from Table~\ref{tab:quant_res}, we qualitatively assess \lossname{}'s performance in Fig.~\ref{fig:app_loss}. We find that \lossname{} outperforms all other loss functions not only quantitatively, but also qualitatively. Compared to optimizing the DiceCE, soft-clDice, and cbDice loss functions, the \methodname{} version trained on \lossname{} is capable of segmenting a more complete and detailed vascular network, given its intrinsic imbalance-aware weighting scheme. The model trained with the Skeleton Recall loss severely over-segments vascular structures and produces a large number of FPs. In contrast, by explicitly penalizing FPs via its FP penalty, \lossname{} yields considerably more accurate and well-localized vessel delineations.

\subsection{Additional Implementation Details}\label{app:impl}
Complementing Section~\ref{sec:impl}, we provide additional implementation details on \methodname{}, the free-text segmentation models, and our synthetic vasculature generation process.

\paragraph{\methodname{}:}
\Methodname{} is trained using the Adam optimizer~\cite{kingma2014adam} with an initial learning rate of $1 \times 10^{-5}$, which is finally decayed to $1 \times 10^{-6}$. We set the batch size to 8 and operate on a patch size of $128 \times 128 \times 128$. \Methodname{}'s architecture relies on an updated configuration. We therefore import the \texttt{ResidualEncoderUNet} architecture from the \texttt{dynamic-network-architectures} package and initialize it with the \texttt{nnUNetResEncUNetL} config derived from nnU-Net. We further apply label smoothing during training, shifting ground truth labels from \{0, 1\} to \{0.025, 0.975\}. \Methodname{} applies a combination of spatial and intensity-based data augmentations during training.
Spatial augmentations include random rotations ($p_\text{rot}=0.2$), zooming ($p_\text{zoom}=0.2$), and flipping ($p_\text{flip}=0.2$ per axis). Intensity augmentations include intensity scaling ($p_\text{scale}=0.2$), intensity shifting ($p_\text{shift}=0.2$), contrast adjustment ($p_\text{contrast}=0.2$), histogram shifting ($p_\text{hist}=0.1$), Gaussian smoothing ($p_\text{smooth}=0.1$), and adding Gaussian noise ($p_\text{noise}=0.1$). Finally, we clip HU values to the range of [-1000, 1000] and subsequently rescale them to [0,1].

\paragraph{Free-Text Universal Medical Segmentation Models (BiomedParse \& VoxTell):}
We use v2 of the BiomedParse model and follow the author's recommended preprocessing steps. For experiments with VoxTell, we load the model checkpoint v1.1.
Final predictions from the free-text segmentation models used in our experiments are obtained by aggregating predictions from 21 prompts covering both vascular structure-specific and more general descriptions. The final segmentation is produced by merging the individual outputs corresponding to each prompt. We observe that this structured prompting strategy substantially improves performance, as single, coarse prompts such as \texttt{"Complete cardiovascular system in CT"} fail to capture the full anatomical complexity. The 21 text prompts are listed below:
\begin{verbatim}
text_prompts = [
    "Aorta in CT", 
    "Inferior vena cava in CT", 
    "Superior vena cava in CT", 
    "Pulmonary artery in CT", 
    "Pulmonary veins in CT",
    "Heart in CT", 
    "Coronary arteries in CT",
    "Common iliac arteries in CT", 
    "Common iliac veins in CT",
    "Portal vein in CT",
    "Carotid arteries in CT", 
    "Subclavian arteries in CT", 
    "Femoral arteries in CT", 
    "Jugular veins in CT",
    "Hepatic vasculature in CT",
    "Pulmonary vasculature in CT",
    "Renal vasculature in CT",
    "Peripheral small-caliber vessels in CT",
    "Systemic vasculature in CT", 
    "All blood vessels in CT", 
    "Complete cardiovascular system in CT"
]
\end{verbatim}

\paragraph{Synthetic Vasculature Generation:}
We utilize synthetic vasculature during the first iteration of our training process (see Section~\ref{sec:training}). We build vascular trees using the \texttt{Forest} class from the \texttt{svv} package. We set \texttt{n\_networks} to 1, \texttt{n\_trees\_per\_network} to 2, \texttt{physical\_clearance} to 0.01 and \texttt{compete} to \texttt{True}. We generate a total of 10,000 vessel segments. For each training image, we generate 10 synthetic samples.

\subsection{Additional Qualitative Results}\label{app:more_qual_res}
\begin{figure}[h]
\centering
\includegraphics[width=\linewidth]{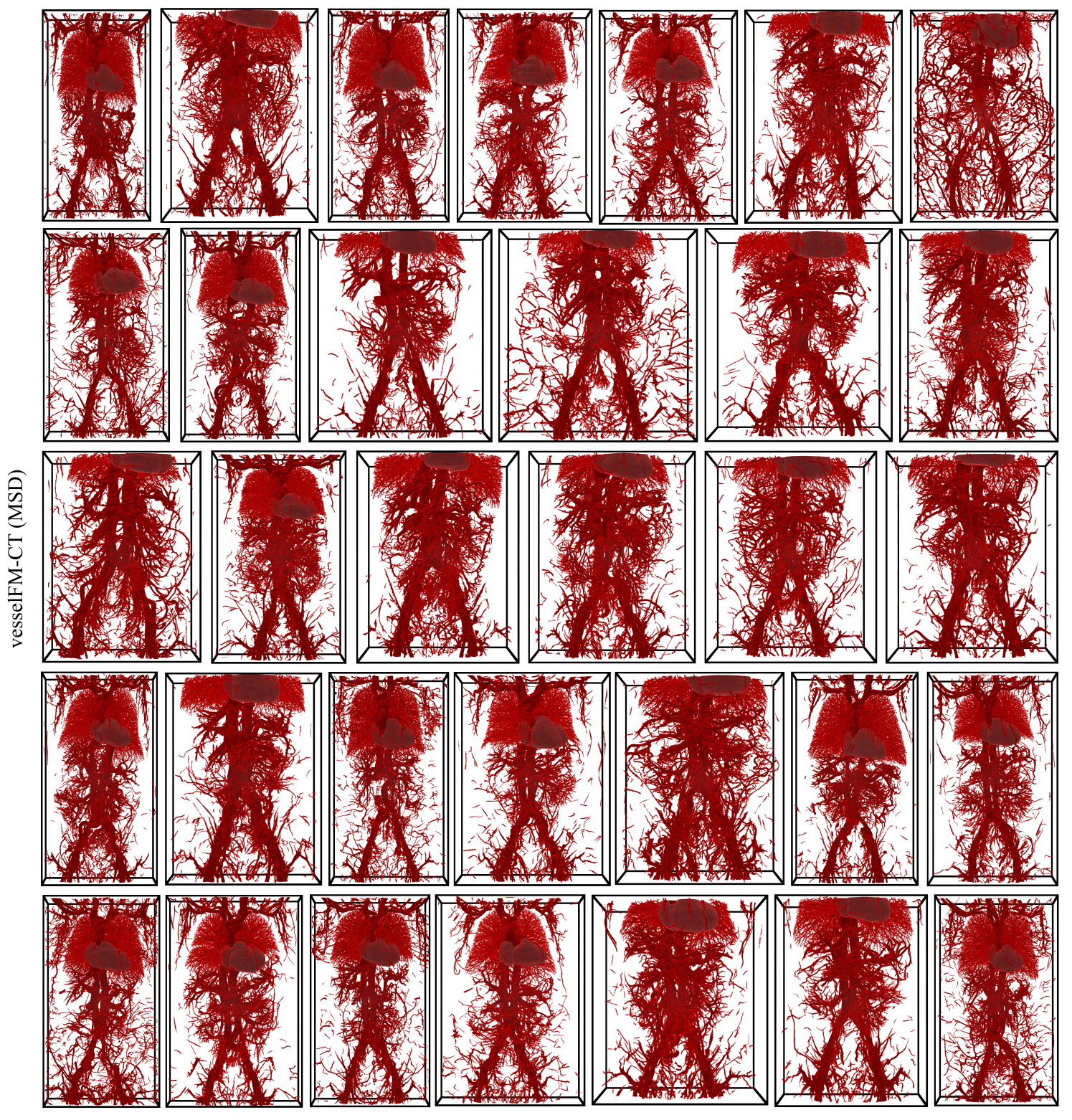}
\vspace{-1em}
\caption{
Additional qualitative results of \methodname{} on the MSD dataset (\texttt{Task03\_Liver}).
}
\label{fig:app_qual_res_msd}
\end{figure}

\begin{figure}[h]
\centering
\includegraphics[width=\linewidth]{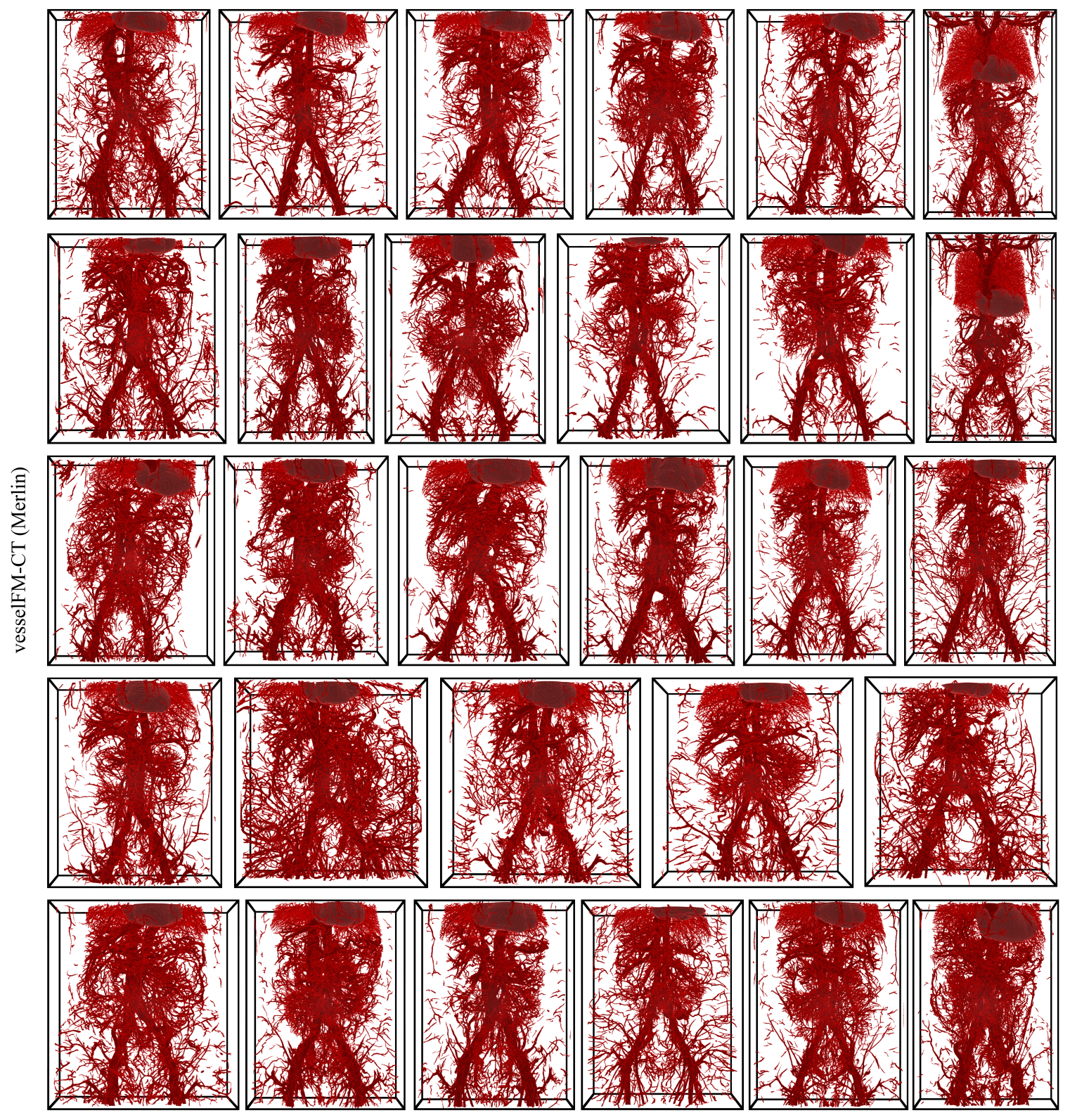}
\vspace{-1em}
\caption{
Additional qualitative results of \methodname{} on CT images from the Merlin dataset.
}
\label{fig:app_qual_res_merlin}
\end{figure}

We present additional qualitative results of \methodname{} on data from MSD \texttt{Task03\_Liver} (see Fig.~\ref{fig:app_qual_res_msd}) and Merlin (see Fig.~\ref{fig:app_qual_res_merlin}). We find that \methodname{} consistently predicts comprehensive segmentation masks of the cardiovascular network. Notably, \methodname{} is robust to domain shifts, segmenting structure from the Merlin dataset in unprecedented detail, although exclusively trained on data from MSD.

\subsection{Limitations} 
\Methodname{} is currently not capable of multi-class cardiovascular segmentation and assigns binary labels. Although all of its components are suited for multi-class segmentation tasks, acquiring accurate multi-class anatomical labels for all substructures of the cardiovascular system represents an immense overhead. Future work should, therefore, aim at extending \methodname{} to support finer-grained anatomical differentiation. Furthermore, \methodname{} is restricted to a single imaging modality, namely CT. Extending the approach to other modalities, such as MRI, would require dedicated model adaptations. Finally, our downstream experiments are conducted as a thorough proof of concept and do not constitute a comprehensive evaluation of all potential clinical and technical applications unlocked by \methodname{}.

\subsection{Additional Information on Downstream Experiments}\label{app:down}   % 3,266 samples
\paragraph{Merlin Subset:} Both downstream experiments are conducted on the Merlin dataset. We subsample the Merlin dataset as described in Section~\ref{sec:down}. We solely include diseases that have or may have an effect on the cardiovascular system to explore the benefits of our \methodname{}-based classifier. We, therefore, exclude the classes: submucosal edema, pancreatic atrophy, renal cyst, osteopenia, surgically absent gallbladder, hiatal hernia, biliary ductal dilation, appendicitis, gallstones, bowel obstruction, free air, and fracture. %Our final subset contains 3,266 image-label pairs.

\subsubsection{Disease Classification}\label{app:dis_cls}   % 3,251 samples
\paragraph{Training Details:} During training, we use a masked binary CE loss, where loss terms corresponding to uncertain labels (label -1) are ignored, and gradients are propagated only for labels indicating definite presence (label 1) or absence (label 0). The learning rate is set to $1 \times 10^{-5}$, the effective batch size to 32 (batch size of 4 and 8 gradient accumulation steps), and the number of training epochs to 200. We divide our Merlin subset into training, validation, and test sets using iterative stratification~\cite{Sechidis2011OnTS} to preserve the multi-label distribution across splits. The resulting splits contain 2,279, 327, and 645 scans, respectively. Both models are of a comparable number of trainable parameters ($\sim$8.2M params.). All experiments are conducted on a single NVIDIA A100 GPU (80 GB VRAM). Training the image-based classifier takes approximately 7 days, whereas the \methodname{}-based sparse classifier converges in approximately 24 hours. This difference is primarily attributed to I/O bottlenecks.
% However, the major bottleneck in image-based classifiers is the data loading. In contrast, the \methodname{}-based classifier can load the entire dataset into memory, significantly accelerating training.

\begin{table}[t]
\centering
\small
\caption{Detailed results of all diseases and conditions in our Merlin subset. We compare the \methodname{}-based classifier (Ours) operating on segmentation masks to the image-based classifier (Baseline).}
\label{tab:quant_res_down_full}
\setlength{\tabcolsep}{4pt}
\begin{tabular}{llccccccc}
\toprule
& \multirow{3}{*}{Disease / Condition}
& \multicolumn{3}{c}{Ours}
&& \multicolumn{3}{c}{Baseline} \\
\cmidrule(lr){3-5} \cmidrule(lr){7-9}
&& F1 & AUC & BA && F1 & AUC & BA \\
\midrule

\multirow{5}{*}{\rotatebox{90}{\textit{cardiovasc.}}}
& Atherosclerosis & 82.79 & 85.08 & 76.73 && \textbf{86.76} & \textbf{90.06} & \textbf{81.65} \\
& Coronary Artery Calcification & \textbf{58.82} & \textbf{76.38} & \textbf{69.82} && 53.57 & 68.97 & 66.30 \\
& Aortic Valve Calcification & \textbf{87.72} & \textbf{92.86} & \textbf{86.64} && 82.76 & 89.43 & 80.86 \\
& Abdominal Aortic Aneurysm & \textbf{58.33} & \textbf{90.47} & \textbf{76.52} && 51.43 & 79.11 & 68.71 \\
& Thrombosis & \textbf{72.73} & \textbf{58.78} & \textbf{57.43} && 63.93 & 50.95 & 53.85 \\
\midrule

\multirow{5}{*}{\rotatebox{90}{\textit{enlarg.}}}
& Cardiomegaly & \textbf{77.55} & \textbf{87.53} & \textbf{84.70} && 60.47 & 80.07 & 71.66 \\
& Hepatomegaly & 43.75 & 74.39 & \textbf{66.26} && \textbf{44.90} & \textbf{77.63} & 66.11 \\
& Splenomegaly & \textbf{38.98} & \textbf{79.89} & \textbf{75.76} && 24.72 & 72.26 & 60.07 \\
& Prostatomegaly & \textbf{34.29} & 70.45 & \textbf{65.11} && 13.79 & \textbf{70.98} & 52.86 \\
& Lymphadenopathy & 45.00 & 56.11 & 58.52 && \textbf{51.85} & \textbf{65.31} & \textbf{61.69} \\
\midrule

\multirow{8}{*}{\rotatebox{90}{\textit{others}}}
& Ascites & 87.34 & 91.55 & 77.39 && \textbf{91.67} & \textbf{95.25} & \textbf{89.47} \\
& Anasarca & 74.87 & 90.06 & 84.18 && \textbf{81.52} & \textbf{93.29} & \textbf{88.65} \\
& Atelectasis & \textbf{80.32} & \textbf{81.84} & \textbf{76.12} && 71.97 & 74.96 & 68.14 \\
& Pleural Effusion & \textbf{89.26} & \textbf{93.87} & \textbf{84.75} && 82.69 & 87.51 & 78.77 \\
& Hepatic Steatosis & 70.83 & 54.20 & 58.69 && \textbf{81.74} & \textbf{67.67} & \textbf{59.96} \\
& Hydronephrosis & \textbf{23.33} & \textbf{63.91} & \textbf{60.85} && 19.47 & 60.59 & 55.28 \\
& Renal Hypodensities & \textbf{41.27} & \textbf{67.12} & \textbf{62.66} && 23.33 & 64.56 & 51.31 \\
& Metastatic Disease & \textbf{51.11} & \textbf{77.61} & \textbf{73.02} && 41.94 & 73.15 & 63.49 \\
\bottomrule
\end{tabular}
\end{table}

\paragraph{Image-based Classifier (Baseline):}
The image-based baseline directly processes the original CT images using a 3D ResNet-18 backbone.
The classifier is implemented using the MONAI~\cite{cardoso2022monai} 3D ResNet-18 architecture. The network uses 3D convolutions, a single input channel, and a feed-forward classification head whose output dimension matches the number of class labels. The initial convolution uses a kernel size of 7 and a stride of 2, followed by max pooling. To reduce memory usage, the channel width is scaled by 0.5. CT intensities are clipped to [-1000, 1000] HU and normalized to [0, 1].

\paragraph{\methodname{}-based Classifier (Ours):}
The \methodname{}-based model operates on the segmentation maps predicted by \methodname{} and augments them with HU values from the corresponding image regions, to retain intensity information. In addition, we include the heart label from the CADS-model for completeness. We replace the standard dense convolutional ResNet backbone with a sparse architecture similar to~\cite{prabhakar2026sparse}. We closely follow the design of the image-based classifier, substituting dense 3D convolutions with sparse convolutional layers.
\paragraph{Additional Quantitative Results:}
We report area under the receiver operating characteristic curve (AUC) on the test set of cardiovascular diseases in Section~\ref{sec:down}. For completeness, we report additionally quantitative results for all diseases and conditions in our Merlin subset in Table~\ref{tab:quant_res_down_full}. We report AUC, along with balanced accuracy (BA), and F1 scores. To compute threshold-dependent metrics, we determine a separate classification threshold for each abnormality using the validation set. Specifically, we select the threshold corresponding to the point on the receiver operating characteristic curve (ROC) closest to the top-left corner, following~\cite{hamamci2026generalist}. Interestingly, we find that the \methodname{}-based classifier performs not only superior on cardiovascular diseases, but also on conditions that have an effect on the cardiovascular system (\eg, metastatic disease (\textbf{77.61} vs. 73.15 AUC) or splenomegaly (\textbf{79.89} vs. 72.26  AUC)).

\subsubsection{Synthetic CT Image Generation}\label{app:ct_gen} % 3180  images
We argue that generating realistic CT images requires accurate delineation of blood vessels. Additionally, sub-millimeter-level blood vessel segmentation could provide long-range context across anatomical regions, leading to a more realistic representation. While segmentation-conditioned synthetic CT generation has achieved high-quality results, generating realistic blood vessels at the submillimeter level in CT images remains an unmet need. To this end, we leverage \methodname{}'s capabilities to segment all vessels and use it to train a conditional CT generation method. For this, we primarily rely on the MAISI-v2 model, which is trained to generate CT images conditioned on organ-level anatomical segmentation masks and further enhance its capabilities to incorporate additional vessel segmentation masks.

\paragraph{Fine-Tuning MAISI-v2 ControlNet:} The MAISI-v2 architecture is based on a denoising U-Net trained with a rectified flow objective and controllable via a ControlNet. We used the pretrained ControlNet from MAISI-v2 and fine-tuned it on the Merlin dataset, keeping the denoising U-Net frozen. For this, we used the CADS-model to segment all anatomical regions of the Merlin datasets. Subsequently, we map the CADS label IDs into MAISI-compatible label IDs. For the experiments, we created two sets of label maps: 1) Only the combined CADS label map to use as a baseline to see if the model learns vessel structure as an emerging property, and 2) we overwrote the combined CADS label maps \methodname{}'s label map whenever there is a blood vessel. For the latter case, we use a 130 value (previously 'dummy') to represent the blood vessel label. We subsequently run two fine-tuning experiments with these two condition sets.

\paragraph{Training Details:} For fine-tuning, we split images from our Merlin subset meeting the size criteria into 2544 train and 636 test cases. The z-dimension was cropped to one of the values in [128, 256, 384, 512, 640]. The model is trained on 4 NVIDIA H100 GPUs (96 GB VRAM) for approximately 2 days, with a batch size of 1 per GPU and the AdamW optimizer with a learning rate of $1 \times 10^{-5}$ for 20 epochs.

\begin{figure}[h]
\centering
\includegraphics[width=0.95\linewidth]{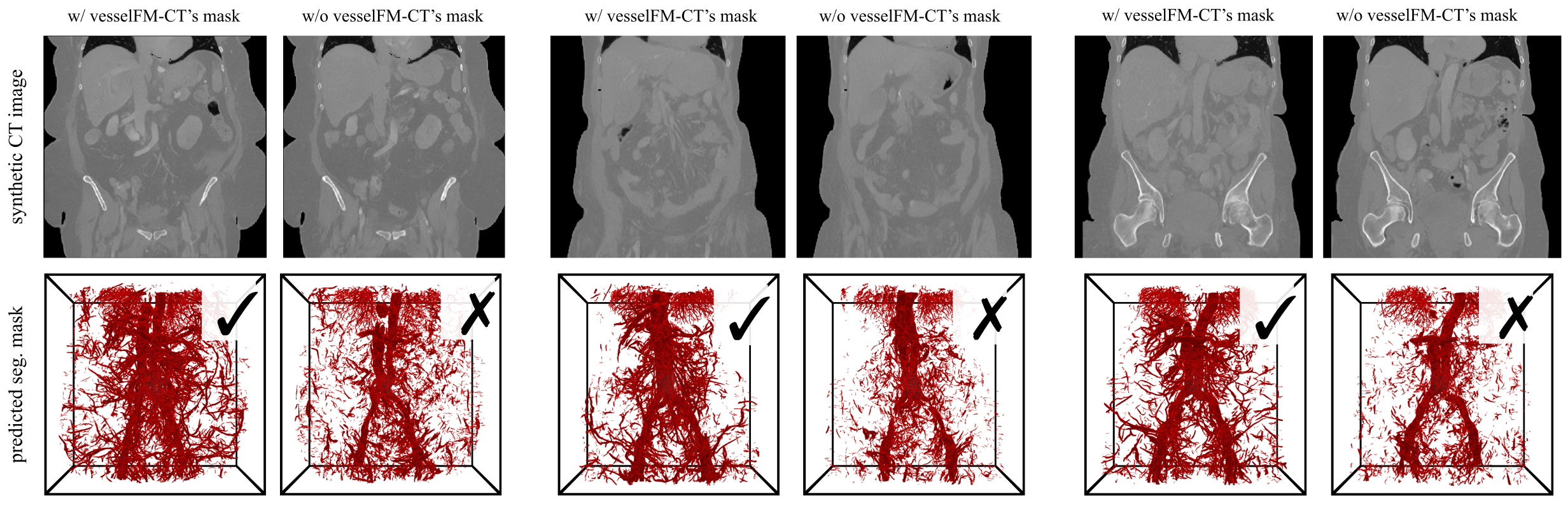}
\vspace{-1em}
\caption{
We show synthetic, generated images in the first row, and predicted segmentation masks, verifying the completeness of the cardiovascular system in the second row. We compare MAISI-v2 variants, where we fine-tuned the ControlNet with and without our generated segmentation masks.  
}
\label{fig:app_qual_res_maisi}
\end{figure}

\paragraph{Additional Qualitative Results:} 
We show additional qualitative results comparing the two MAISI-v2 variants in Fig.~\ref{fig:app_qual_res_maisi}. We find that incorporating vesselFM-CT's predicted mask during ControlNet fine-tuning yields superior synthetic images that better preserve cardiovascular structure.

% \newpage
% \input{checklist.tex}

\end{document}